# LEARNING IN A LARGE FUNCTION SPACE: PRIVACY-PRESERVING MECHANISMS FOR SVM LEARNING

BENJAMIN I. P. RUBINSTEIN[1], PETER L. BARTLETT[1,2], LING HUANG[3], NINA TAFT[3]
[1]Computer Science Division, UC Berkeley [2]Department of Statistics, UC Berkeley [3]Intel Labs Berkeley

ABSTRACT. Several recent studies in privacy-preserving learning have considered the trade-off between utility or risk and the level of differential privacy guaranteed by mechanisms for statistical query processing. In this paper we study this trade-off in private Support Vector Machine (SVM) learning. We present two efficient mechanisms, one for the case of finite-dimensional feature mappings and one for potentially infinite-dimensional feature mappings with translation-invariant kernels. For the case of translation-invariant kernels, the proposed mechanism minimizes regularized empirical risk in a random Reproducing Kernel Hilbert Space whose kernel uniformly approximates the desired kernel with high probability. This technique, borrowed from large-scale learning, allows the mechanism to respond with a finite encoding of the classifier, even when the function class is of infinite VC dimension. Differential privacy is established using a proof technique from algorithmic stability. Utility—the mechanism's response function is pointwise $\epsilon$-close to non-private SVM with probability $1-\delta$—is proven by appealing to the smoothness of regularized empirical risk minimization with respect to small perturbations to the feature mapping. We conclude with a lower bound on the optimal differential privacy of the SVM. This negative result states that for any $\delta$, no mechanism can be simultaneously $(\epsilon, \delta)$-useful and $\beta$-differentially private for small $\epsilon$ and small $\beta$.

## 1. INTRODUCTION

The goal of a well-designed statistical database is to provide aggregate information about a database's entries while maintaining individual entries' privacy. These two goals of *utility* and *privacy* are inherently discordant. For a mechanism to be useful, its responses must closely resemble some target statistic of the database's entries. However to protect privacy, it is often necessary for the mechanism's response distribution to be 'smoothed out', *i.e.*, the mechanism must be randomized to reduce the individual entries' influence on this distribution. It has been of key interest to the statistical database community to understand when the goals of utility and privacy can be efficiently achieved simultaneously (Dinur and Nissim, 2003; Barak et al., 2007; Dwork et al., 2007; Blum et al., 2008; Chaudhuri and Monteleoni, 2009; Kasiviswanathan et al., 2008). In this paper we consider the practical goal of private regularized empirical risk minimization (ERM) in Reproducing Kernel Hilbert Spaces for the special case of the Support Vector Machine (SVM). We adopt the strong notion of differential privacy as formalized by Dwork (2006). Our efficient new mechanisms are shown to parametrize functions that are close to non-private SVM under the $L_\infty$-norm, with high probability. In our setting this notion of utility is stronger than closeness of risk (*cf.* Remark 3).

We employ a number of algorithmic and proof techniques new to differential privacy. One of our new mechanisms borrows a technique from large-scale learning, in which regularized ERM is performed in a random feature space whose inner-product uniformly approximates the target feature space inner-product. This random feature space is constructed by viewing the target kernel as a probability measure in the Fourier domain. This technique enables the finite parametrization of responses from function classes with infinite VC dimension. To establish utility, we show that regularized ERM is relatively insensitive to perturbations of the kernel: not only does the technique of learning in a random RKHS enable finitely-encoded privacy-preserving responses, but these responses well-approximate the responses of non-private SVM. Together these two techniques may prove useful in extending privacy-preserving mechanisms to learn in large function spaces. To prove differential privacy, we borrow a proof technique from the area of algorithmic stability. We believe that stability may become a fruitful avenue for constructing new private mechanisms in the future, based on learning maps presently known to be stable.

Of particular interest, is the optimal differential privacy of the SVM, which loosely speaking is the best level of privacy achievable by any accurate mechanism for SVM learning. Through our privacy-preserving mechanisms for the SVM, endowed with guarantees of utility, we upper bound optimal differential privacy. We also provide lower bounds on the SVM's optimal differential privacy, which are impossibility results for simultaneously achieving high levels of utility and privacy.

The remainder of this paper is organized as follows. After concluding this section with a summary of related work, we recall basic concepts of differential privacy and SVM learning in Section 2. Sections 3 and 4 describe the new mechanisms for private SVM learning for finite-dimensional feature maps and (potentially infinite-dimensional) feature maps with translation-invariant kernels. Each mechanism is accompanied with proofs of privacy and utility bounds. Section 5 considers the special case of hinge loss and presents an upper bound on the SVM's optimal differential privacy. A corresponding lower bound is then given in Section 6. We conclude the paper with several open problems.

1.1. **Related Work.** There is a rich literature of prior work on differential privacy in the theory community. The following sections summarize work related to our own, organized to contrast this work with our main contributions.

1.1.1. *Range Spaces Parametrizing Vector-Valued Statistics or Functions with Finite VC-dimension.* Early work on private interactive mechanisms focused on approximating real- and vector-valued statistics (*e.g.*, Dinur and Nissim, 2003; Blum et al., 2005; Dwork et al., 2006; Dwork, 2006; Barak et al., 2007). McSherry and Talwar (2007) first considered private mechanisms with range spaces parametrizing sets more general than real-valued vectors, and used such differentially private mappings for mechanism design. More related to our work are the private mechanisms for regularized logistic regression proposed and analyzed by Chaudhuri and Monteleoni (2009). There the mechanism's range space parametrizes the VC-dimension $d+1$ class of linear hyperplanes in $\mathbb{R}^d$. Kasiviswanathan et al. (2008) showed that discretized concept classes can be PAC learned or agnostically learned privately, albeit via an inefficient mechanism. Blum et al. (2008) showed that non-interactive mechanisms can privately release anonymized data such that utility is guaranteed over classes of predicate queries with polynomial VC dimension, when the domain is discretized. Dwork et al. (2009) more recently characterized when utility and privacy can be achieved by efficient non-interactive mechanisms. In this paper we consider efficient mechanisms for private SVM learning, whose range spaces parametrize real-valued functions (whose sign form trained classifiers). One case covered by our analysis is learning with a Gaussian kernel, which corresponds to learning over a class of infinite VC dimension.

1.1.2. *Practical Privacy-Preserving Learning (Mostly) via Subset-Sums.* Most prior work in differential privacy has focused on the deep analysis of mechanisms for relatively simple statistics (with histograms and contingency tables as explored by Blum et al. 2005 and Barak et al. 2007 respectively, as examples) and learning algorithms (*e.g.*, interval queries and half-spaces as explored by Blum et al. 2008), or on constructing learning algorithms that can be decomposed into subset-sum operations (*e.g.*, perceptron, $k$-NN, ID3 as described by Blum et al. 2005, and various recommender systems due to the work of McSherry and Mironov 2009). By contrast, we consider the practical goal of SVM learning, which does not decompose into subset-sums. It is also notable that our mechanisms run in polynomial time. The most related work to our own in this regard is due to Chaudhuri and Monteleoni (2009), although their results hold only for differentiable loss, and finite feature mappings.

1.1.3. *The Privacy-Utility Trade-Off.* Like several prior studies, we consider the trade-off between privacy and utility. Barak et al. (2007) presented a mechanism for releasing contingency tables that guarantees differential privacy and also guarantees a notion of accuracy: with high probability all marginals from the released table are close in $L_1$-norm to the true table's marginals. As mentioned above, Blum et al. (2008) developed a private non-interactive mechanism that releases anonymized data such that all predicate queries in a VC-class take on similar values on the anonymized data and original data. In the work of Kasiviswanathan et al. (2008), utility corresponds to PAC learning: with high probability the response and target concepts are close, averaged over the underlying measure.

A sequence of prior negative results have shown that any mechanism providing overly accurate responses cannot be private (Dinur and Nissim, 2003; Dwork et al., 2007; Dwork and Yekhanin, 2008). Dinur and Nissim (2003) showed that if noise of rate only $o(\sqrt{n})$ is added to subset sum queries on a database of bits then an adversary can reconstruct a $1-o(1)$ fraction of the database. This is a threshold phenomenon that says if accuracy is too small, privacy cannot be guaranteed at all. This result was more recently extended to allow for mechanisms that answer a small fraction of queries arbitrarily (Dwork et al., 2007). We show a similar negative result for the private SVM setting: any mechanism that is too accurate with respect to the SVM cannot guarantee strong levels of privacy.

1.1.4. *Connections between Stability, Robust Statistics, and Global Sensitivity.* To prove differential privacy, we borrow a proof technique from the area of algorithmic stability. In passing Kasiviswanathan et al. (2008) note the similarity between notions of algorithmic stability and differential privacy, however do not exploit this. The connection between algorithmic stability and differential privacy is qualitatively similar to the recent work of Dwork and Lei (2009) who demonstrated that robust estimators can serve as the basis for private mechanisms, by exploiting the limited influence of outliers on such estimators.

## 2. Background & Definitions

A *database* $D$ is a sequence of $n > 1$ *entries* or *rows* $(\mathbf{x}_i, y_i) \in \mathbb{R}^d \times \{-1, 1\}$, which are input point-label pairs or *examples*. We say that a pair of databases $D_1, D_2$ are *neighbors* if they differ on one entry. A *mechanism* $M$ is a service trusted with access to a database $D$, that releases aggregate information about $D$ while maintaining privacy of individual entries. By $M(D)$ we mean the *response* of $M$ on $D$. We assume that this is the only information released by the mechanism. Denote the range space of $M$ by $\mathcal{T}_M$. We adopt the following strong notion of differential privacy due to Dwork (2006).

**Definition 1.** For any $\beta > 0$, a randomized mechanism $M$ provides *$\beta$-differential privacy*, if, for all neighboring databases $D_1, D_2$ and all responses $t \in \mathcal{T}_M$ ,

$$\log\left(\frac{\Pr\left(M(D_1) = t\right)}{\Pr\left(M(D_2) = t\right)}\right) \quad \leq \quad \beta \, .$$

The probability in the definition is over the randomization in $M$. For continuous $\mathcal{T}_M$ we mean by this ratio a Radon-Nikodym derivative of the distribution of $M(D_1)$ with respect to the distribution of $M(D_2)$. If an adversary knows $M$ and the first $n-1$ entries of $D$, she may simulate the mechanism with different choices for the missing example. If the mechanism's response distribution varies smoothly with her choice, the adversary will not be able to infer the true value of entry $n$ by querying $M$. In the sequel we assume WLOG that each pair of neighboring databases differ on their last entry.

Intuitively the more an 'interesting'[1] mechanism $M$ is perturbed to guarantee differential privacy, the less like $M$ the resulting mechanism $\hat{M}$ will become. The next definition formalizes the notion of 'likeness'.

**Definition 2.** Consider two mechanisms $\hat{M}$ and $M$ with the same domain and response spaces $\mathcal{T}_{\hat{M}}, \mathcal{T}_M$ respectively. Let $\mathcal{X}$ be some set and let $\mathcal{F}$ be a space of real-valued functions on $\mathcal{X}$ that is parametrized by the response spaces: for every $t \in \mathcal{T}_{\hat{M}} \cup \mathcal{T}_M$ let $f_t \in \mathcal{F}$ be some function. Finally assume $\mathcal{F}$ is endowed with norm $\|\cdot\|_{\mathcal{F}}$. Then for $\epsilon > 0$ and $0 < \delta < 1$ we say that[2] $\hat{M}$ is *$(\epsilon, \delta)$-useful* with respect to $M$ if, for all databases $D$, $\Pr\left(\left\|f_{\hat{M}(D)} - f_{M(D)}\right\|_{\mathcal{F}} \leq \epsilon\right) \geq \delta$.

Typically $\hat{M}$ will be a privacy-preserving version of $M$, that has been perturbed somehow. Usefulness means that not only does $\hat{M}$ guarantee privacy of the training database, but that the aggregate information revealed about the database by $\hat{M}$ is 'close' to what would be revealed by the desired (but non-private) mechanism $M$. In the sequel we will take $\|\cdot\|_{\mathcal{F}}$ to be the sup-norm over a subset $\mathcal{M} \subseteq \mathbb{R}^d$ containing the data, which we denote by $\|f\|_{\infty;\mathcal{M}} = \sup_{\mathbf{x}\in\mathcal{M}} |f(\mathbf{x})|$. It will also be convenient to use the notation $\|k\|_{\infty;\mathcal{M}} = \sup_{\mathbf{x},\mathbf{y}\in\mathcal{M}} |k(\mathbf{x}, \mathbf{y})|$ for bivariate functions $k(\cdot, \cdot)$.

*Remark* 3. In the sequel we develop privacy-preserving mechanisms that are useful with respect to the Support Vector Machine (see the next section for a brief introduction to the SVM). The SVM works to minimize the expected hinge-loss (*i.e.*, risk in terms of the hinge-loss), which is a convex surrogate for the expected 0-1 loss. Since the hinge-loss is Lipschitz in the real-valued function output by the SVM, it follows that a mechanism $\hat{M}$ having utility with respect to the SVM also has expected hinge-loss that is within

[1] Examples of interesting properties include low risk, robustness to a small amount of malicious noise, etc.

[2] Note that we have chosen to overload the term $(\epsilon, \delta)$-usefulness introduced by Blum et al. (2008) for non-interactive mechanisms that release anonymized data. Our definition of usefulness is analogous for the present setting of privacy-preserving learning, where a single function is released.

**Algorithm 1** SVM

**Inputs:** database $D = \{(\mathbf{x}_i, y_i)\}_{i=1}^n$ with $\mathbf{x}_i \in \mathbb{R}^d$, $y_i \in \{-1, 1\}$; kernel $k : \mathbb{R}^d \times \mathbb{R}^d \to \mathbb{R}$; convex loss function $\ell$; parameter $C > 0$.

(1) $\boldsymbol{\alpha}^\star \leftarrow$ Solve the QP dual of Primal (2.1) (see *e.g.*, the derivations by Bishop 2006); and
(2) Return vector $\boldsymbol{\alpha}^\star$.

$\epsilon$ of the SVM's hinge-loss with high probability. That is, $(\epsilon, \delta)$-usefulness with respect to the sup-norm is stronger than guaranteed closeness of risk (absolute bounds on risk for regularized logistic regression are explored by Chaudhuri and Monteleoni 2009; Kasiviswanathan et al. 2008 consider the task of private PAC learning, which demands closeness of risk). We consider the hinge-loss further in Sections 5 and 6. Until then we work with arbitrary convex, Lipschitz losses.

We will see that the presented analysis does not simultaneously guarantee privacy at arbitrary levels and utility at arbitrary accuracy. The highest level of privacy guaranteed over all $(\epsilon, \delta)$-useful mechanisms with respect to a target mechanism $M$, is quantified by the optimal differential privacy for $M$. We define this notion for the SVM here, but the concept extends to any target mechanism of interest. We present upper and lower bounds on $\beta(\epsilon, \delta, C, n, \ell, k)$ for the SVM in Sections 5 and 6 respectively.

**Definition 4.** For $\epsilon, C > 0$, $\delta \in (0, 1)$, $n > 1$, loss function $\ell(y, \hat{y})$ convex in $\hat{y}$, and kernel $k$, the *optimal differential privacy for the SVM* is the function

$$\beta(\epsilon, \delta, C, n, \ell, k) \quad = \quad \inf_{\hat{M} \in \mathcal{I}} \sup_{(D_1, D_2) \in \mathcal{D}} \sup_{t \in \mathcal{T}_{\hat{M}}} \log \left( \frac{\Pr\left(\hat{M}(D_1) = t\right)}{\Pr\left(\hat{M}(D_2) = t\right)} \right) ,$$

where $\mathcal{I}$ is the set of all $(\epsilon, \delta)$-useful mechanisms with respect to the SVM with parameter $C$, loss $\ell$, and kernel $k$; and $\mathcal{D}$ is the set of all pairs of neighboring databases with $n$ entries.

2.1. **Background on Support Vector Machines.** Soft-margin SVM learning corresponds to the convex Primal program

$$\min_{\mathbf{w} \in \mathbb{R}^F} \quad \frac{1}{2}\|\mathbf{w}\|_2^2 + \frac{C}{n} \sum_{i=1}^n \ell\left(y_i, f_{\mathbf{w}}(\mathbf{x}_i)\right) \ , \tag{2.1}$$

where the $\mathbf{x}_i \in \mathbb{R}^d$ are *training input points* and the $y_i \in \{-1, 1\}$ are their *training labels*, $n$ is the size of the training set, $\phi : \mathbb{R}^d \to \mathbb{R}^F$ is a *feature mapping* taking points in *input space* $\mathbb{R}^d$ to some (possibly infinite) $F$-dimensional *feature space*, $\ell(y, \hat{y})$ is a loss function convex in $\hat{y}$, and $\mathbf{w}$ is a hyperplane *normal vector* in feature space.

When $F$ is finite, predictions are made by taking the sign of $f^\star(\mathbf{x}) = f_{\mathbf{w}^\star}(\mathbf{x}) = \langle \phi(\mathbf{x}), \mathbf{w}^\star \rangle$. We will refer to both $f_{\mathbf{w}}(\cdot)$ and $\operatorname{sgn}(f_{\mathbf{w}}(\cdot))$ as *classifiers*, with the exact meaning apparent from the context. When $F$ is large and when inner-products in feature space may be computed quickly via an explicit representation of the *kernel function* $k(\mathbf{x}, \mathbf{y}) = \langle \phi(\mathbf{x}), \phi(\mathbf{y}) \rangle$, the solution may be more easily obtained via the dual. For example, see Program (5.1) in Section 5 for the dual formulation of the hinge-loss $\ell(y, \hat{y}) = (1 - y\hat{y})_+$, which is the loss most commonly associated with soft-margin SVM. Other examples include the square loss $(1 - y\hat{y})^2$ and logistic loss $\log(1 + \exp(-y\hat{y}))$. The vector of maximizing dual variables $\boldsymbol{\alpha}^\star$ returned by dualized SVM parametrizes the function $f^\star = f_{\boldsymbol{\alpha}^\star}$ as $f_{\boldsymbol{\alpha}}(\cdot) = \sum_{i=1}^m \alpha_i y_i k(\cdot, \mathbf{x}_i)$.

More generally, the Support Vector Machine can be seen as performing regularized ERM in a Reproducing Kernel Hilbert Space (RKHS) $\mathcal{H}$. The Representer Theorem (Kimeldorf and Wahba, 1971) states that the minimizing $f^\star = \arg\min_{f \in \mathcal{H}} \frac{1}{2}\|f\|_{\mathcal{H}}^2 + \frac{C}{n} \sum_{i=1}^n \ell(y_i, f(\mathbf{x}_i))$ lies in the span of the functions $k(\cdot, \mathbf{x}_i) \in \mathcal{H}$. Indeed the above dual expansion shows that the coordinates in this subspace are given by the $\alpha_i^\star y_i$.

We define the *mechanism* SVM to be the dual optimization that responds with the vector $\boldsymbol{\alpha}^\star$, as described by Algorithm 1. For general information about SVMs see *e.g.*, (Burges, 1998; Cristianini and Shawe-Taylor, 2000; Schölkopf and Smola, 2001; Bishop, 2006). We end this section with the definition of an important class of kernels (see Table 1 for examples).

**Definition 5.** A kernel function of the form $k(\mathbf{x}, \mathbf{y}) = g(\mathbf{x} - \mathbf{y})$, for some function $g$, is called *translation-invariant*.

| *Kernel* | $g(\Delta)$ | $p(\omega)$ |
|---|---|---|
| RBF | $\exp\left(-\frac{\|\Delta\|_2^2}{2\sigma^2}\right)$ | $(2\pi)^{-d/2}\exp\left(-\frac{\|\omega\|_2^2}{2}\right)$ |
| Laplacian | $\exp\left(-\|\Delta\|_1\right)$ | $\prod_{i=1}^d \frac{1}{\pi\left(1+\omega_i^2\right)}$ |
| Cauchy | $\prod_{i=1}^d \frac{2}{1+\Delta_i^2}$ | $\exp\left(-\|\Delta\|_1\right)$ |

TABLE 1. Example translation-invariant kernels, their $g$ functions and the corresponding Fourier transforms.

## 3. MECHANISM FOR FINITE FEATURE MAPS

As a first step towards private SVM learning we begin by considering the simple case of finite $F$-dimensional feature maps. Algorithm 2 describes the PRIVATESVM-FINITE mechanism, which follows the usual pattern of preserving differential privacy: after forming the primal solution to the SVM—an $F$-dimensional vector—the mechanism adds Laplace-distributed noise to the weight vector. Guaranteeing differential privacy proceeds via the usual two-step process of calculating the $L_1$-sensitivity of the SVM's weight vector, then showing that $\beta$-differential privacy follows from sensitivity together with the choice of Laplace noise with scale equal to sensitivity divided by $\beta$.

To calculate sensitivity, we exploit the algorithmic stability of regularized ERM. Intuitively, stability corresponds to continuity of a learning map. Several notions of stability are known to lead to good generalization error bounds (Devroye and Wagner, 1979; Kearns and Ron, 1999; Bousquet and Elisseeff, 2002; Kutin and Niyogi, 2002), sometimes in cases where class capacity-based approaches such as VC theory do not apply. A *learning map* $\mathcal{A}$ is a function that maps a database $D$ to a classifier $f_D$; it is precisely the composition of a mechanism followed by the classifier parametrization mapping.[3] A learning map $\mathcal{A}$ is said to have *$\gamma$-uniform stability* with respect to loss $\ell(\cdot,\cdot)$ if for all neighboring databases $D, D'$, the losses of the classifiers trained on $D$ and $D'$ are close on all test examples $\|\ell(\cdot,\mathcal{A}(D)) - \ell(\cdot,\mathcal{A}(D'))\|_\infty \le \gamma$ (Bousquet and Elisseeff, 2002). Our first lemma computes sensitivity by following the proof of (Schölkopf and Smola, 2001, Theorem 12.4) which establishes that SVM learning has uniform stability (a result due to Bousquet and Elisseeff 2002). For simplicity we restrict the proof of sensitivity to differentiable loss functions in Lemma 6; the result remains the same for general convex loss functions. See Lemma 21 for an almost identical proof for subdifferentiable losses.

**Lemma 6.** *Consider loss function $\ell(y,\hat{y})$ that is differentiable, convex and L-Lipschitz in $\hat{y}$, and an RKHS $\mathcal{H}$ induced by finite $F$-dimensional feature mapping $\phi$ with bounded norm $k(\mathbf{x},\mathbf{x}) \le \kappa^2$ for all $\mathbf{x} \in \mathbb{R}^d$. Let $\mathbf{w}_S \in \mathbb{R}^F$ be the minimizer of the following regularized empirical risk function for each database $S = \{(\mathbf{x}_i, y_i)\}_{i=1}^n$*

$$R_{\text{reg}}(\mathbf{w}, S) \quad = \quad \frac{C}{n}\sum_{i=1}^n \ell\left(y_i, f_{\mathbf{w}}(\mathbf{x}_i)\right) + \frac{1}{2}\|\mathbf{w}\|_2^2 \ .$$

*Then for every pair of neighboring databases $D, D'$ of $n$ entries, $\|\mathbf{w}_D - \mathbf{w}_{D'}\|_1 \le 4LC\kappa\sqrt{F}/n$.*

*Proof.* For convenience we define $R_{\text{emp}}(\mathbf{w}, S) = n^{-1}\sum_{i=1}^n \ell\left(y_i, f_{\mathbf{w}}(\mathbf{x}_i)\right)$ for any database $S$, then the first-order necessary KKT conditions imply

$$\partial_{\mathbf{w}} R_{\text{reg}}(\mathbf{w}_D, D) \quad = \quad C\partial_{\mathbf{w}} R_{\text{emp}}(\mathbf{w}_D, D) + \mathbf{w}_D \ = \ \mathbf{0} \tag{3.1}$$

$$\partial_{\mathbf{w}} R_{\text{reg}}(\mathbf{w}_{D'}, D') \quad = \quad C\partial_{\mathbf{w}} R_{\text{emp}}(\mathbf{w}_{D'}, D') + \mathbf{w}_{D'} \ = \ \mathbf{0} \ , \tag{3.2}$$

where $\partial_{\mathbf{w}}$ is the partial derivative operator with respect to $\mathbf{w}$. Define the auxiliary risk function

$$\tilde{R}(\mathbf{w}) \quad = \quad C\langle \partial_{\mathbf{w}} R_{\text{emp}}(\mathbf{w}_D, D) - \partial_{\mathbf{w}} R_{\text{emp}}(\mathbf{w}_{D'}, D'),\, \mathbf{w} - \mathbf{w}_{D'}\rangle + \frac{1}{2}\|\mathbf{w} - \mathbf{w}_{D'}\|_2^2 \ .$$

It is easy to see that $\tilde{R}(\mathbf{w})$ is strictly convex in $\mathbf{w}$ and that $\tilde{R}(\mathbf{w}_{D'}) = 0$. And since by Equation (3.2)

$$\begin{aligned} \partial_{\mathbf{w}} \tilde{R}(\mathbf{w}) \quad &= \quad C\partial_{\mathbf{w}} R_{\text{emp}}(\mathbf{w}_D, D) - C\partial_{\mathbf{w}} R_{\text{emp}}(\mathbf{w}_{D'}, D') + \mathbf{w} - \mathbf{w}_{D'} \\ &= \quad C\partial_{\mathbf{w}} R_{\text{emp}}(\mathbf{w}_D, D) + \mathbf{w} \ , \end{aligned}$$

[3]For example an SVM mechanism may return a weight vector $\mathbf{w}^\star$ or dual coefficients $\boldsymbol{\alpha}^\star$ which in turn parametrizes the classifier $f^\star$. The SVM *learning map* takes the training database directly to the classifier.

**Algorithm 2** PrivateSVM-Finite

**Inputs:** database $D = \{(\mathbf{x}_i, y_i)\}_{i=1}^n$ with $\mathbf{x}_i \in \mathbb{R}^d$, $y_i \in \{-1, 1\}$; finite feature map $\phi : \mathbb{R}^d \to \mathbb{R}^F$ and induced kernel $k$; convex loss function $\ell$; and parameters $\lambda, C > 0$.

(1) $\boldsymbol{\alpha}^\star \leftarrow$ Run Algorithm 1 on $D$ with parameter $C$, kernel $k$, and loss $\ell$;
(2) $\tilde{\mathbf{w}} \leftarrow \sum_{i=1}^n \alpha_i^\star y_i \phi(\mathbf{x}_i)$;
(3) $\boldsymbol{\mu} \leftarrow$ Draw i.i.d. sample of $F$ scalars from Laplace$(0, \lambda)$; and
(4) Return $\hat{\mathbf{w}} = \tilde{\mathbf{w}} + \boldsymbol{\mu}$

it follows that $\tilde{R}(\mathbf{w})$ is minimized at $\mathbf{w}_D$ by Equation (3.1). Thus $\tilde{R}(\mathbf{w}_D) \leq 0$. Next simplify the first term of $\tilde{R}(\mathbf{w}_D)$, scaled by $n/C$ for simplicity:

$$\begin{aligned}
& n\langle \partial_{\mathbf{w}} R_{\text{emp}}(\mathbf{w}_D, D) - \partial_{\mathbf{w}} R_{\text{emp}}(\mathbf{w}_{D'}, D'),\, \mathbf{w}_D - \mathbf{w}_{D'} \rangle \\
= \; & \sum_{i=1}^n \langle \partial_{\mathbf{w}} \ell(y_i, f_{\mathbf{w}_D}(\mathbf{x}_i)) - \partial_{\mathbf{w}} \ell(y_i', f_{\mathbf{w}_{D'}}(\mathbf{x}_i')),\, \mathbf{w}_D - \mathbf{w}_{D'} \rangle \\
= \; & \sum_{i=1}^{n-1} \left(\ell'(y_i, f_{\mathbf{w}_D}(\mathbf{x}_i)) - \ell'(y_i, f_{\mathbf{w}_{D'}}(\mathbf{x}_i))\right)\left(f_{\mathbf{w}_D}(\mathbf{x}_i) - f_{\mathbf{w}_{D'}}(\mathbf{x}_i)\right) \\
& + \ell'(y_n, f_{\mathbf{w}_D}(\mathbf{x}_n))\left(f_{\mathbf{w}_D}(\mathbf{x}_n) - f_{\mathbf{w}_{D'}}(\mathbf{x}_n)\right) - \ell'(y_n', f_{\mathbf{w}_{D'}}(\mathbf{x}_n'))\left(f_{\mathbf{w}_D}(\mathbf{x}_n') - f_{\mathbf{w}_{D'}}(\mathbf{x}_n')\right) \\
\geq \; & \ell'(y_n, f_{\mathbf{w}_D}(\mathbf{x}_n))\left(f_{\mathbf{w}_D}(\mathbf{x}_n) - f_{\mathbf{w}_{D'}}(\mathbf{x}_n)\right) - \ell'(y_n', f_{\mathbf{w}_{D'}}(\mathbf{x}_n'))\left(f_{\mathbf{w}_D}(\mathbf{x}_n') - f_{\mathbf{w}_{D'}}(\mathbf{x}_n')\right) ,
\end{aligned}$$

where $\ell'(y, \hat{y}) = \partial_{\hat{y}} \ell(y, \hat{y})$. The second equality follows from $\partial_{\mathbf{w}} \ell(y, f_{\mathbf{w}}(\mathbf{x})) = \ell'(y, f_{\mathbf{w}}(\mathbf{x}))\,\phi(\mathbf{x})$ and $\mathbf{x}_i' = \mathbf{x}_i$ and $y_i' = y_i$ for each $i \in [n-1]$, and the inequality follows from the differentiability and convexity[4] of $\ell$ in $\hat{y}$. Combined with $\tilde{R}(\mathbf{w}_D) \leq 0$ this yields

$$\begin{aligned}
& \frac{n}{2C}\|\mathbf{w}_D - \mathbf{w}_{D'}\|_2^2 \\
\leq \; & \ell'(y_n', f_{\mathbf{w}_{D'}}(\mathbf{x}_n'))\left(f_{\mathbf{w}_D}(\mathbf{x}_n') - f_{\mathbf{w}_{D'}}(\mathbf{x}_n')\right) - \ell'(y_n, f_{\mathbf{w}_D}(\mathbf{x}_n))\left(f_{\mathbf{w}_D}(\mathbf{x}_n) - f_{\mathbf{w}_{D'}}(\mathbf{x}_n)\right) \\
\leq \; & 2L\left\|f_{\mathbf{w}_D} - f_{\mathbf{w}_{D'}}\right\|_\infty , \qquad (3.3)
\end{aligned}$$

by the Lipschitz continuity of $\ell$. Now by the reproducing property and Cauchy-Schwartz inequality we can upper bound the classifier difference's infinity norm by the Euclidean norm on the weight vectors: for each $\mathbf{x}$

$$\begin{aligned}
\left|f_{\mathbf{w}_D}(\mathbf{x}) - f_{\mathbf{w}_{D'}}(\mathbf{x})\right| &= |\langle \phi(\mathbf{x}), \mathbf{w}_D - \mathbf{w}_{D'} \rangle| \\
&\leq \|\phi(\mathbf{x})\|_2 \|\mathbf{w}_D - \mathbf{w}_{D'}\|_2 \\
&= \sqrt{k(\mathbf{x}, \mathbf{x})}\, \|\mathbf{w}_D - \mathbf{w}_{D'}\|_2 \\
&\leq \kappa \|\mathbf{w}_D - \mathbf{w}_{D'}\|_2 .
\end{aligned}$$

Combining this with Inequality (3.3) yields $\|\mathbf{w}_D - \mathbf{w}_{D'}\|_2 \leq 4LC\kappa/n$. $L_1$-based sensitivity then follows from the inequality $\|\mathbf{w}\|_1 \leq \sqrt{F}\|\mathbf{w}\|_2$ for all $\mathbf{w} \in \mathbb{R}^F$. □

With the weight vector's sensitivity in hand, differential privacy follows immediately from the proof technique established by Dwork et al. (2006).

**Theorem 7** (Privacy of PrivateSVM-Finite)**.** *For any $\beta > 0$, database $D$ of size $n$, $C > 0$, loss function $\ell(y, \hat{y})$ that is convex and $L$-Lipschitz in $\hat{y}$, and finite $F$-dimensional feature map with kernel $k(\mathbf{x}, \mathbf{x}) \leq \kappa^2$ for all $\mathbf{x} \in \mathbb{R}^d$, PrivateSVM-Finite run on $D$ with loss $\ell$, kernel $k$, noise parameter $\lambda \geq 4LC\kappa\sqrt{F}/(\beta n)$ and regularization parameter $C$ guarantees $\beta$-differential privacy.*

This first main result establishes the usual kind of differential privacy guarantee for the new PrivateSVM-Finite algorithm. The more "private" the data, the more noise must be added. The more entries in the database, the less noise is needed to achieve the same level of privacy. Since the noise vector $\boldsymbol{\mu}$ has exponential tails, standard tail bound inequalities quickly lead to $(\epsilon, \delta)$-usefulness for PrivateSVM-Finite.

[4]Namely for differentiable convex $f$ and any $a, b \in \mathbb{R}$, $(f'(a) - f'(b))(a - b) \geq 0$.

**Theorem 8** (Utility of PrivateSVM-Finite)**.** *Consider any $C > 0$, $n > 1$, database $D$ of $n$ entries, arbitrary convex loss $\ell$, and finite $F$-dimensional feature mapping $\phi$ with kernel $k$ and $|\phi(\mathbf{x})_i| \leq \Phi$ for all $\mathbf{x} \in \mathcal{M}$ and $i \in [F]$ for some $\Phi > 0$ and $\mathcal{M} \subseteq \mathbb{R}^d$. For any $\epsilon > 0$ and $\delta \in (0, 1)$,* PrivateSVM-Finite *run on $D$ with loss $\ell$, kernel $k$, noise parameter $0 < \lambda \leq \frac{\epsilon}{2\Phi\left(F \log_e 2 + \log_e \frac{1}{\delta}\right)}$, and regularization parameter $C$, is $(\epsilon, \delta)$-useful with respect to the* SVM *under the $\|\cdot\|_{\infty;\mathcal{M}}$-norm.*

*Proof.* Our goal is to compare the SVM and PrivateSVM-Finite classifications of any point $\mathbf{x} \in \mathcal{M}$:

$$\begin{aligned} \left| f_{\hat{M}(D)}(\mathbf{x}) - f_{M(D)}(\mathbf{x}) \right| &= |\langle \hat{\mathbf{w}}, \phi(\mathbf{x}) \rangle - \langle \tilde{\mathbf{w}}, \phi(\mathbf{x}) \rangle| \\ &= |\langle \boldsymbol{\mu}, \phi(\mathbf{x}) \rangle| \\ &\leq \|\boldsymbol{\mu}\|_1 \, \|\phi(\mathbf{x})\|_\infty \\ &\leq \Phi \, \|\boldsymbol{\mu}\|_1 \ . \end{aligned}$$

The absolute value of a zero mean Laplace random variable with scale parameter $\lambda$ is exponentially distributed with scale $\lambda^{-1}$. Moreover the sum of $q$ i.i.d. exponential random variables has Erlang $q$-distribution with the same scale parameter.[5] Thus we have, for Erlang $F$-distributed random variable $X$ and any $t > 0$,

$$\begin{aligned} \forall \mathbf{x} \in \mathcal{M}, \ \left| f_{\hat{M}(D)}(\mathbf{x}) - f_{M(D)}(\mathbf{x}) \right| &\leq \Phi X \\ \Rightarrow \ \forall \epsilon > 0, \ \Pr\left( \left\| f_{\hat{M}(D)} - f_{M(D)} \right\|_{\infty;\mathcal{M}} > \epsilon \right) &\leq \Pr\left( X > \epsilon / \Phi \right) \\ &= \Pr\left( e^{tX} > e^{t\epsilon/\Phi} \right) \\ &\leq \frac{\mathbb{E}\left[ e^{tX} \right]}{e^{\epsilon t/\Phi}} \ . \end{aligned} \tag{3.4}$$

Here we have employed the standard Chernoff tail bound technique using Markov's inequality. The numerator of (3.4), the moment generating function of the Erlang $F$-distribution with parameter $\lambda$, is $(1 - \lambda t)^{-F}$ for all $t < \lambda^{-1}$. Together with the choice of $t = (2\lambda)^{-1}$, this gives

$$\begin{aligned} \Pr\left( \left\| f_{\hat{M}(D)} - f_{M(D)} \right\|_{\infty;\mathcal{M}} > \epsilon \right) &\leq (1 - \lambda t)^{-F} \, e^{-\epsilon t/\Phi} \\ &= 2^F e^{-\epsilon/(2\lambda\Phi)} \\ &= \exp\left( F \log_e 2 - \epsilon/(2\lambda\Phi) \right) \ . \end{aligned}$$

And provided that $\lambda \leq \epsilon / \left( 2\Phi \left( F \log_e 2 + \log_e \frac{1}{\delta} \right) \right)$ this probability is bounded by $\delta$. ☐

Our second main result establishes that PrivateSVM-Finite is not only differentially private, but that it releases a classifier that is similar to the SVM. Utility and privacy are competing properties, however, since utility demands that the noise not be too large.

## 4. Mechanism for Translation-Invariant Kernels

Consider now the problem of privately learning in an RKHS $\mathcal{H}$ induced by an infinite dimensional feature mapping $\phi$. As a mechanism's response must be finitely encodable, the primal parametrization seems less appealing as it did in PrivateSVM-Finite. It is natural to look to the SVM's dual solution as a starting point: the Representer Theorem (Kimeldorf and Wahba, 1971) states that the optimizing $f^\star \in \mathcal{H}$ must be in the span of the data—a finite-dimensional subspace. While the coordinates in this subspace—the $\alpha_i^\star$ dual variables—could be perturbed in the usual way to guarantee differential privacy, the subspace's basis—the data—are also needed to parametrize $f^\star$. To side-step this apparent stumbling block, we take another approach by approximating $\mathcal{H}$ with a random RKHS $\hat{\mathcal{H}}$ induced by a random finite-dimensional map $\hat{\phi}$. This then allows us to respond with a finite primal parametrization. Algorithm 3 summarizes the PrivateSVM mechanism.

As noted recently by Rahimi and Recht (2008), the Fourier transform $p$ of the $g$ function of a continuous positive-definite translation-invariant kernel is a non-negative measure (Rudin, 1994). Rahimi and Recht

[5]The Erlang $q$-distribution has density $\frac{x^{q-1}\exp(-x/\lambda)}{\lambda^q (q-1)!}$, CDF $1 - e^{-x/\lambda} \sum_{j=0}^{q-1} \frac{(x/\lambda)^j}{j!}$, expectation $q\lambda$ and variance $q\lambda^2$.

**Algorithm 3** PrivateSVM

**Inputs:** database $D = \{(\mathbf{x}_i, y_i)\}_{i=1}^n$ with $\mathbf{x}_i \in \mathbb{R}^d$, $y_i \in \{-1, 1\}$; translation-invariant kernel $k(\mathbf{x}, \mathbf{y}) = g(\mathbf{x} - \mathbf{y})$ with Fourier transform $p(\boldsymbol{\omega}) = 2^{-1} \int e^{-j\langle \boldsymbol{\omega}, \mathbf{x} \rangle} g(\mathbf{x})\, d\mathbf{x}$; convex loss function $\ell$; parameters $\lambda, C > 0$ and $\hat{d} \in \mathbb{N}$.

(1) $\boldsymbol{\rho}_1, \ldots, \boldsymbol{\rho}_{\hat{d}} \leftarrow$Draw i.i.d. sample of $\hat{d}$ vectors in $\mathbb{R}^d$ from $p$;
(2) $\hat{\boldsymbol{\alpha}} \leftarrow$ Run Algorithm 1 on $D$ with parameter $C$, kernel $\hat{k}$ induced by map (4.1), and loss $\ell$;
(3) $\tilde{\mathbf{w}} \leftarrow \sum_{i=1}^n y_i \hat{\alpha}_i \hat{\phi}(\mathbf{x}_i)$ where $\hat{\phi}$ is defined in Equation (4.1);
(4) $\boldsymbol{\mu} \leftarrow$ Draw i.i.d. sample of $2\hat{d}$ scalars from Laplace $(0, \lambda)$; and
(5) Return $\hat{\mathbf{w}} = \tilde{\mathbf{w}} + \boldsymbol{\mu}$ and $\boldsymbol{\rho}_1, \ldots, \boldsymbol{\rho}_{\hat{d}}$

(2008) exploit this fact to construct a random finite-dimensional RKHS $\hat{\mathcal{H}}$ by drawing $\hat{d}$ vectors from $p$. These vectors $\boldsymbol{\rho}_1, \ldots, \boldsymbol{\rho}_{\hat{d}}$ define the following random $2\hat{d}$-dimensional feature map

$$\hat{\phi}(\cdot) \quad = \quad \hat{d}^{-1/2} \left[\cos\left(\langle \boldsymbol{\rho}_1, \cdot \rangle\right), \sin\left(\langle \boldsymbol{\rho}_1, \cdot \rangle\right), \ldots, \cos\left(\langle \boldsymbol{\rho}_{\hat{d}}, \cdot \rangle\right), \sin\left(\langle \boldsymbol{\rho}_{\hat{d}}, \cdot \rangle\right)\right]^T \ . \tag{4.1}$$

Inner-products in the random feature space approximate $k(\cdot, \cdot)$ uniformly, and to arbitrary precision depending on parameter $\hat{d}$, as restated in Lemma 13. We denote the inner-product in the random feature space by $\hat{k}$. Rahimi and Recht (2008) applied this approximation to large-scale learning (situations where $n$ is large). Instead of employing non-linear SVM's dual solution which takes $O(n^2)$ time, the primal solution to linear SVM on $\hat{\phi}$ is used, as it takes time quadratic in $\hat{d}$ to compute. For large-scale learning, good approximations can be found for $\hat{d} \ll n$. Table 1 presents three important translation-invariant kernels and their transformations. PrivateSVM employs the same trick for translation-invariant kernels, but in a different setting. Here regularized ERM is performed in $\hat{\mathcal{H}}$, not to avoid complexity in $n$, but to provide a direct finite representation $\tilde{\mathbf{w}}$ of the primal solution in the case of infinite dimensional feature spaces. After performing regularized ERM in $\hat{\mathcal{H}}$, appropriate Laplace noise is added to the primal solution $\tilde{\mathbf{w}}$ to guarantee differential privacy as before.

PrivateSVM is computationally efficient. Algorithm 3 takes $O(\hat{d})$ time to compute each entry of the kernel matrix, or a total time of $O(\hat{d}n^2)$ on top of running dual SVM in the random feature space which is worst-case $O(n_s^3)$ for the analytic solution (where $n_s \le n$ is the number of support vectors), and faster using numerical methods such as chunking (Burges, 1998). To achieve $(\epsilon, \delta)$-usefulness wrt the hinge-loss SVM $\hat{d}$ must be taken to be $O\left(\frac{d}{\epsilon^4}\left(\log\frac{1}{\delta} + \log\frac{1}{\epsilon}\right)\right)$ (*cf.* Corollary 15). By comparison it takes $O(dn^2)$ to construct the kernel matrix for any translation-invariant kernel.

As with the SVM and PrivateSVM-Finite, the response of Algorithm 3 can be used to make classifications on future test points by constructing the classifier $\hat{f}^\star(\cdot) = f_{\hat{\mathbf{w}}}(\cdot) = \langle \hat{\mathbf{w}}, \hat{\phi}(\cdot) \rangle$. Unlike the previous mechanisms, however, PrivateSVM must include a parametrization of feature map $\hat{\phi}$—the sample $\{\boldsymbol{\rho}_i\}_{i=1}^{\hat{d}}$—in its response. Of PrivateSVM's total response, only $\hat{\mathbf{w}}$ depends on database $D$. The $\boldsymbol{\rho}_i$ are data-independent vectors drawn from the transform $p$ of the kernel, which we assume to be known by the adversary (to wit the adversary knows the mechanism itself, including $k$). Thus to establish differential privacy we need only consider the data-dependent weight vector, fortunately we have already considered the similar case of PrivateSVM-Finite.

**Corollary 9** (Privacy of PrivateSVM). *For any $\beta > 0$, database $D$ of size $n$, $C > 0$, $\hat{d} \in \mathbb{N}$, loss function $\ell(y, \hat{y})$ that is convex and $L$-Lipschitz in $\hat{y}$, and translation-invariant kernel $k$,* PrivateSVM *run on $D$ with loss $\ell$, kernel $k$, noise parameter $\lambda \ge 2^{2.5} LC\sqrt{\hat{d}}/(\beta n)$, approximation parameter $\hat{d}$, and regularization parameter $C$ guarantees $\beta$-differential privacy.*

*Proof.* The result follows immediately from Theorem 7 since $\tilde{\mathbf{w}}$ is the primal solution of SVM with kernel $\hat{k}$, the response vector $\hat{\mathbf{w}} = \tilde{\mathbf{w}} + \boldsymbol{\mu}$ for i.i.d. Laplace $\boldsymbol{\mu}$, and $\hat{k}(\mathbf{x}, \mathbf{x}) = 1$ for all $\mathbf{x} \in \mathbb{R}^D$. ☐

This result is surprising, in that PrivateSVM is able to guarantee privacy for regularized ERM over a function class of infinite VC-dimension, where the obvious way to return the learned classifier (responding with the dual variables and feature mapping) reveals all the entries corresponding to the support vectors, completely.

Like PrivateSVM-Finite, PrivateSVM is useful with respect to the SVM. If we denote the function parametrized by intermediate weight vector $\tilde{\mathbf{w}}$ by $\tilde{f}$, then the same argument for the utility of PrivateSVM-Finite establishes the high-probability proximity of $\tilde{f}$ and $f^\star$.

**Lemma 10.** *Consider a run of Algorithms 1 and 3 with $\hat{d} \in \mathbb{N}$, $C > 0$, convex loss and translation-invariant kernel. Denote by $\hat{f}^\star$ and $\tilde{f}$ the classifiers parametrized by weight vectors $\hat{\mathbf{w}}$ and $\tilde{\mathbf{w}}$ respectively, where these vectors are related by $\hat{\mathbf{w}} = \tilde{\mathbf{w}} + \boldsymbol{\mu}$ with $\boldsymbol{\mu} \overset{iid}{\sim} \mathrm{Laplace}(0, \lambda)$ in Algorithm 3. For any $\epsilon > 0$ and $\delta \in (0, 1)$, if $0 < \lambda \le \min\left\{\frac{\epsilon}{2^4 \log_e 2\sqrt{\hat{d}}}, \frac{\epsilon\sqrt{\hat{d}}}{8\log_e \frac{2}{\delta}}\right\}$ then*

$$\Pr\left(\left\|\hat{f}^\star - \tilde{f}\right\|_\infty \le \frac{\epsilon}{2}\right) \quad \ge \quad 1 - \frac{\delta}{2}\ .$$

*Proof.* As in the proof of Theorem 8 we can use the Chernoff trick to show that, for Erlang $2\hat{d}$-distributed random variable $X$, the choice of $t = (2\lambda)^{-1}$,and for any $\epsilon > 0$

$$\begin{aligned}\Pr\left(\left\|\hat{f}^\star - \tilde{f}\right\|_\infty > \epsilon/2\right) &\le \frac{\mathbb{E}\left[e^{tX}\right]}{e^{\epsilon t\sqrt{\hat{d}}/2}} \\ &\le (1-\lambda t)^{-2\hat{d}} e^{-\epsilon t\sqrt{\hat{d}}/2} \\ &= 2^{2\hat{d}} e^{-\epsilon\sqrt{\hat{d}}/(4\lambda)} \\ &= \exp\left(\hat{d}\log_e 4 - \epsilon\sqrt{\hat{d}}/(4\lambda)\right)\ .\end{aligned}$$

Provided that $\lambda \le \epsilon / \left(2^4 \log_e 2\sqrt{\hat{d}}\right)$ this is bounded by $\exp\left(-\epsilon\sqrt{\hat{d}}/(8\lambda)\right)$. Moreover if $\lambda \le \epsilon\sqrt{\hat{d}} / \left(8\log_e \frac{2}{\delta}\right)$, then the claim follows. □

To show a similar result for $f^\star$ and $\tilde{f}$, we exploit smoothness of the regularized ERM with respect to small changes in the RKHS itself. To the best of our knowledge, this kind of stability to the feature mapping has not been used before. We begin with a technical lemma that we will use to exploit the convexity of the regularized empirical risk functional.

**Lemma 11.** *Let $R$ be a functional on Hilbert space $\mathcal{H}$ satisfying $R[f] \ge R[f^\star] + \frac{a}{2}\|f - f^\star\|_{\mathcal{H}}^2$ for some $a > 0$, $f^\star \in \mathcal{H}$ and all $f \in \mathcal{H}$. Then $R[f] \le R[f^\star] + \epsilon$ implies $\|f - f^\star\|_{\hat{\mathcal{H}}} \le \sqrt{\frac{2\epsilon}{a}}$, for all $\epsilon > 0$, $f \in \mathcal{H}$.*

*Proof.* By assumption and the antecedent

$$\begin{aligned}\|f - f^\star\|_{\hat{\mathcal{H}}}^2 &\le \frac{2}{a}\left(R[f] - R[f^\star]\right) \\ &\le \frac{2}{a}\left(R[f^\star] + \epsilon - R[f^\star]\right) \\ &= 2\epsilon/a\ .\end{aligned}$$

Taking square roots of both sides yields the consequent. □

Provided that the kernel functions $k$ and $\hat{k}$ are uniformly close, the next lemma exploits insensitivity of regularized ERM to perturbations of the feature mapping to show that $f^\star$ and $\tilde{f}$ are pointwise close. Lemma 22 re-proves this result for non-differentiable loss functions.

**Lemma 12.** *Let $\mathcal{H}$ be an RKHS with translation-invariant kernel $k$, and let $\hat{\mathcal{H}}$ be the random RKHS corresponding to feature map (4.1) induced by $k$. Let $C$ be a positive scalar and loss $\ell(y, \hat{y})$ be differentiable, convex, and L-Lipschitz in $\hat{y}$. Consider the regularized empirical risk minimizers in each RKHS*

$$\begin{aligned}f^\star &\in \arg\min_{f\in\mathcal{H}} \frac{C}{n}\sum_{i=1}^n \ell(y_i, f(\mathbf{x}_i)) + \frac{1}{2}\|f\|_{\mathcal{H}}^2\ , \\ g^\star &\in \arg\min_{g\in\hat{\mathcal{H}}} \frac{C}{n}\sum_{i=1}^n \ell(y_i, g(\mathbf{x}_i)) + \frac{1}{2}\|g\|_{\hat{\mathcal{H}}}^2\ .\end{aligned}$$

*Let $\mathcal{M} \subseteq \mathbb{R}^d$ be any set containing $\mathbf{x}_1, \ldots, \mathbf{x}_n$. For any $\epsilon > 0$, if the dual variables from both optimizations have $L_1$-norms bounded by some $\Lambda > 0$ and $\left\|k - \hat{k}\right\|_{\infty;\mathcal{M}} \leq \min\left\{1, \frac{\epsilon^2}{2^2\left(\Lambda+2\sqrt{(CL+\Lambda/2)\Lambda}\right)^2}\right\}$ then $\|f^\star - g^\star\|_{\infty;\mathcal{M}} \leq \epsilon/2$.*

*Proof.* Denote the empirical risk functional by $R_{\mathrm{emp}}[f] = n^{-1}\sum_{i=1}^n \ell\left(y_i, f(\mathbf{x}_i)\right)$ and the regularized empirical risk functional $R_{\mathrm{reg}}[f] = C\, R_{\mathrm{emp}}[f] + \|f\|^2/2$, for the appropriate RKHS norm (either $\mathcal{H}$ or $\hat{\mathcal{H}}$). Let $f^\star$ denote the regularized empirical risk minimizer in $\mathcal{H}$, given by parameter vector $\boldsymbol{\alpha}^\star$, and let $g^\star$ denote the regularized empirical risk minimizer in $\hat{\mathcal{H}}$ given by parameter vector $\boldsymbol{\beta}^\star$. Let $g_{\boldsymbol{\alpha}^\star} = \sum_{i=1}^n \alpha_i^\star y_i \hat{\phi}(\mathbf{x}_i)$ and $f_{\boldsymbol{\beta}^\star} = \sum_{i=1}^n \beta_i^\star y_i \phi(\mathbf{x}_i)$ denote the images of $f^\star$ and $g^\star$ under the natural mapping between the spans of the data in RKHS's $\hat{\mathcal{H}}$ and $\mathcal{H}$ respectively. We will first show that these four functions have arbitrarily close regularized empirical risk in their respective RKHS, and then that this implies uniform proximity of the functions themselves. First observe that for any $g \in \hat{\mathcal{H}}$

$$
\begin{aligned}
R_{\mathrm{reg}}^{\hat{\mathcal{H}}}[g] &= C\, R_{\mathrm{emp}}[g] + \frac{1}{2}\|g\|_{\hat{\mathcal{H}}}^2 \\
&\geq C\langle \partial_g R_{\mathrm{emp}}[g^\star], g - g^\star\rangle_{\hat{\mathcal{H}}} + C\, R_{\mathrm{emp}}[g^\star] + \frac{1}{2}\|g\|_{\hat{\mathcal{H}}}^2 \\
&= \langle \partial_g R_{\mathrm{reg}}^{\hat{\mathcal{H}}}[g^\star], g - g^\star\rangle_{\hat{\mathcal{H}}} - \langle g^\star, g - g^\star\rangle_{\hat{\mathcal{H}}} + C\, R_{\mathrm{emp}}[g^\star] + \frac{1}{2}\|g\|_{\hat{\mathcal{H}}}^2 \\
&= C\, R_{\mathrm{emp}}[g^\star] + \frac{1}{2}\|g\|_{\hat{\mathcal{H}}}^2 - \langle g^\star, g - g^\star\rangle_{\hat{\mathcal{H}}} \\
&= C\, R_{\mathrm{emp}}[g^\star] + \frac{1}{2}\|g^\star\|_{\hat{\mathcal{H}}}^2 + \frac{1}{2}\|g\|_{\hat{\mathcal{H}}}^2 - \frac{1}{2}\|g^\star\|_{\hat{\mathcal{H}}}^2 - \langle g^\star, g - g^\star\rangle_{\hat{\mathcal{H}}} \\
&= R_{\mathrm{reg}}^{\hat{\mathcal{H}}}[g^\star] + \frac{1}{2}\|g\|_{\hat{\mathcal{H}}}^2 - \langle g^\star, g\rangle_{\hat{\mathcal{H}}} + \frac{1}{2}\|g^\star\|_{\hat{\mathcal{H}}}^2 \\
&= R_{\mathrm{reg}}^{\hat{\mathcal{H}}}[g^\star] + \frac{1}{2}\|g - g^\star\|_{\hat{\mathcal{H}}}^2 ,
\end{aligned}
$$

The inequality follows from the convexity of $R_{\mathrm{emp}}[\cdot]$; the subsequent equality by $\partial_g R_{\mathrm{reg}}^{\hat{\mathcal{H}}}[g] = C\,\partial_g R_{\mathrm{emp}}[g] + g$; the third equality by $\partial_g R_{\mathrm{reg}}^{\hat{\mathcal{H}}}[g^\star] = \mathbf{0}$; and the remainder by gathering terms. With this, Lemma 11 states that for any $g \in \hat{\mathcal{H}}$ and $\epsilon' > 0$,

$$
R_{\mathrm{reg}}^{\hat{\mathcal{H}}}[g] \leq R_{\mathrm{reg}}^{\hat{\mathcal{H}}}[g^\star] + \epsilon' \quad \Rightarrow \quad \|g - g^\star\|_{\hat{\mathcal{H}}} \leq \sqrt{2\epsilon'}\ . \tag{4.2}
$$

Next we will show that the antecedent is true for $g = g_{\boldsymbol{\alpha}^\star}$. Conditioned on $\left\{\left\|k - \hat{k}\right\|_{\infty;\mathcal{M}} \leq \epsilon'\right\}$, for all $\mathbf{x} \in \mathcal{M}$

$$
\begin{aligned}
|f^\star(\mathbf{x}) - g_{\boldsymbol{\alpha}^\star}(\mathbf{x})| &= \left|\sum_{i=1}^n \alpha_i^\star y_i \left(k(\mathbf{x}_i, \mathbf{x}) - \hat{k}(\mathbf{x}_i, \mathbf{x})\right)\right| \\
&\leq \sum_{i=1}^n |\alpha_i^\star| \left|k(\mathbf{x}_i, \mathbf{x}) - \hat{k}(\mathbf{x}_i, \mathbf{x})\right| \\
&\leq \epsilon' \|\boldsymbol{\alpha}^\star\|_1 \\
&\leq \epsilon' \Lambda\ , \qquad (4.3)
\end{aligned}
$$

by the bound on $\|\boldsymbol{\alpha}^\star\|_1$. This and the Lipschitz continuity of the loss leads to

$$\begin{aligned}
\left|R_{\mathrm{reg}}^{\mathcal{H}}[f^\star] - R_{\mathrm{reg}}^{\hat{\mathcal{H}}}[g_{\boldsymbol{\alpha}^\star}]\right| &= \left|C\,R_{\mathrm{emp}}[f^\star] - C\,R_{\mathrm{emp}}[g_{\boldsymbol{\alpha}^\star}] + \frac{1}{2}\|f^\star\|_{\mathcal{H}}^2 - \frac{1}{2}\|g_{\boldsymbol{\alpha}^\star}\|_{\hat{\mathcal{H}}}^2\right| \\
&\le \frac{C}{n}\sum_{i=1}^n \left|\ell\left(y_i, f^\star(\mathbf{x}_i)\right) - \ell\left(y_i, g_{\boldsymbol{\alpha}^\star}(\mathbf{x}_i)\right)\right| + \frac{1}{2}\left|\boldsymbol{\alpha}^{\star\prime}\left(\mathbf{K} - \hat{\mathbf{K}}\right)\boldsymbol{\alpha}^\star\right| \\
&\le \frac{C}{n}\sum_{i=1}^n L\,\|f^\star - g_{\boldsymbol{\alpha}^\star}\|_{\infty;\mathcal{M}} + \frac{1}{2}\left|\boldsymbol{\alpha}^{\star\prime}\left(\mathbf{K} - \hat{\mathbf{K}}\right)\boldsymbol{\alpha}^\star\right| \\
&\le CL\,\|f^\star - g_{\boldsymbol{\alpha}^\star}\|_{\infty;\mathcal{M}} + \frac{1}{2}\,\|\boldsymbol{\alpha}^\star\|_1\left\|\left(\mathbf{K} - \hat{\mathbf{K}}\right)\boldsymbol{\alpha}^\star\right\|_\infty \\
&\le CL\,\|f^\star - g_{\boldsymbol{\alpha}^\star}\|_{\infty;\mathcal{M}} + \frac{1}{2}\,\|\boldsymbol{\alpha}^\star\|_1^2\,\epsilon' \\
&\le CL\epsilon'\Lambda + \Lambda^2\epsilon'/2 \\
&= \left(CL + \frac{\Lambda}{2}\right)\Lambda\epsilon' \ .
\end{aligned}$$

Similarly, $\left|R_{\mathrm{reg}}^{\hat{\mathcal{H}}}[g^\star] - R_{\mathrm{reg}}^{\mathcal{H}}[f_{\boldsymbol{\beta}^\star}]\right| \le (CL+\Lambda/2)\Lambda\epsilon'$ by the same argument. And since $R_{\mathrm{reg}}^{\mathcal{H}}[f_{\boldsymbol{\beta}^\star}] \ge R_{\mathrm{reg}}^{\mathcal{H}}[f^\star]$ and $R_{\mathrm{reg}}^{\hat{\mathcal{H}}}[g_{\boldsymbol{\alpha}^\star}] \ge R_{\mathrm{reg}}^{\hat{\mathcal{H}}}[g^\star]$ we have proved that $R_{\mathrm{reg}}^{\hat{\mathcal{H}}}[g_{\boldsymbol{\alpha}^\star}] \le R_{\mathrm{reg}}^{\mathcal{H}}[f^\star] + (CL+\Lambda/2)\Lambda\epsilon' \le R_{\mathrm{reg}}^{\mathcal{H}}[f_{\boldsymbol{\beta}^\star}] + (CL+\Lambda/2)\Lambda\epsilon' \le R_{\mathrm{reg}}^{\hat{\mathcal{H}}}[g^\star] + 2(CL+\Lambda/2)\Lambda\epsilon'$. And by implication (4.2),

$$\|g_{\boldsymbol{\alpha}^\star} - g^\star\|_{\hat{\mathcal{H}}} \ \le\ 2\sqrt{\left(CL + \frac{\Lambda}{2}\right)\Lambda\epsilon'} \ . \tag{4.4}$$

Now $\hat{k}(\mathbf{x}, \mathbf{x}) = 1$ for each $\mathbf{x} \in \mathbb{R}^d$ implies

$$\begin{aligned}
|g_{\boldsymbol{\alpha}^\star}(\mathbf{x}) - g^\star(\mathbf{x})| &= \left\langle g_{\boldsymbol{\alpha}^\star} - g^\star, \hat{k}(\mathbf{x}, \cdot)\right\rangle_{\hat{\mathcal{H}}} \\
&\le \|g_{\boldsymbol{\alpha}^\star} - g^\star\|_{\hat{\mathcal{H}}}\sqrt{\hat{k}(\mathbf{x}, \mathbf{x})} \\
&= \|g_{\boldsymbol{\alpha}^\star} - g^\star\|_{\hat{\mathcal{H}}} \ ,
\end{aligned}$$

This combines with Inequality (4.4) to yield

$$\|g_{\boldsymbol{\alpha}^\star} - g^\star\|_{\infty;\mathcal{M}} \ \le\ 2\sqrt{\left(CL + \frac{\Lambda}{2}\right)\Lambda\epsilon'} \ .$$

Together with Inequality (4.3) this finally implies that $\|f^\star - g^\star\|_{\infty;\mathcal{M}} \le \epsilon'\Lambda + 2\sqrt{(CL+\Lambda/2)\,\Lambda\epsilon'}$, conditioned on event $A_{\epsilon'} = \left\{\left\|k - \hat{k}\right\|_\infty \le \epsilon'\right\}$. For desired accuracy $\epsilon > 0$, conditioning on event $A_{\epsilon'}$ with $\epsilon' = \min\left\{\epsilon/\left[2\left(\Lambda + 2\sqrt{(CL+\Lambda/2)\,\Lambda}\right)\right], \epsilon^2/\left[2\left(\Lambda + 2\sqrt{(CL+\Lambda/2)\,\Lambda}\right)\right]^2\right\}$ yields bound $\|f^\star - g^\star\|_{\infty;\mathcal{M}} \le \epsilon/2$: if $\epsilon' \le 1$ then $\epsilon/2 \ge \sqrt{\epsilon'}\left(\Lambda + 2\sqrt{(CL+\Lambda/2)\,\Lambda}\right) \ge \epsilon'\Lambda + 2\sqrt{(CL+\Lambda/2)\,\Lambda\epsilon'}$ provided that $\epsilon' \le \epsilon^2/\left[2\left(\Lambda + 2\sqrt{(CL+\Lambda/2)\,\Lambda}\right)\right]^2$. Otherwise if $\epsilon' > 1$ then we have $\epsilon/2 \ge \epsilon'\left(\Lambda + 2\sqrt{(CL+\Lambda/2)\,\Lambda}\right) \ge \epsilon'\Lambda + 2\sqrt{(CL+\Lambda/2)\,\Lambda\epsilon'}$ provided $\epsilon' \le \epsilon/\left[2\left(\Lambda + 2\sqrt{(CL+\Lambda/2)\,\Lambda}\right)\right]$. Since for any $H > 0$, $\min\left\{H, H^2\right\} \ge \min\left\{1, H^2\right\}$, the result follows. □

We now recall the result due to Rahimi and Recht (2008) that establishes the non-asymptotic uniform convergence of the kernel functions required by the previous Lemma (*i.e.*, an upper bound on the probability of event $A_{\epsilon'}$).

**Lemma 13** (Rahimi and Recht 2008, Claim 1)**.** *For any $\epsilon > 0$, $\delta \in (0,1)$, translation-invariant kernel $k$ and compact set $\mathcal{M} \subset \mathbb{R}^d$, if $\hat{d} \ge \frac{4(d+2)}{\epsilon^2}\log_e\left(\frac{2^8(\sigma_p \mathrm{diam}(\mathcal{M}))^2}{\delta\epsilon^2}\right)$, then Algorithm 3's random feature mapping $\hat{\phi}$* *defined in Equation* (4.1) *satisfies* $\Pr\left(\left\|\hat{k}-k\right\|_\infty<\epsilon\right)\geq 1-\delta$, *where* $\sigma_p^2=\mathbb{E}\left[\langle\boldsymbol{\omega},\boldsymbol{\omega}\rangle\right]$ *is the second moment of the Fourier transform* $p$ *of* $k$*'s* $g$ *function.*

Combining these ingredients establishes utility for PrivateSVM.

**Theorem 14** (Utility of PrivateSVM)**.** *Consider any database* $D$*, compact set* $\mathcal{M}\subset\mathbb{R}^d$ *containing* $D$*, convex loss* $\ell$*, translation-invariant kernel* $k$*, and scalars* $C,\epsilon>0$ *and* $\delta\in(0,1)$*. Suppose the SVM with loss* $\ell$*, kernel* $k$ *and parameter* $C$ *has dual variables with* $L_1$*-norm bounded by* $\Lambda$*. Then Algorithm 3 run on* $D$ *with loss* $\ell$*, kernel* $k$*, parameters* $\hat{d}\geq\frac{4(d+2)}{\theta(\epsilon)}\log_e\left(\frac{2^9(\sigma_p\mathrm{diam}(\mathcal{M}))^2}{\delta\theta(\epsilon)}\right)$ *where* $\theta(\epsilon)=\min\left\{1,\frac{\epsilon^4}{2^4\left(\Lambda+2\sqrt{(CL+\Lambda/2)\Lambda}\right)^4}\right\}$, $\lambda\leq\min\left\{\frac{\epsilon}{2^4\log_e 2\sqrt{\hat{d}}},\frac{\epsilon\sqrt{\hat{d}}}{8\log_e\frac{2}{\delta}}\right\}$ *and* $C$ *is* $(\epsilon,\delta)$*-useful with respect to Algorithm 1 run on* $D$ *with loss* $\ell$*, kernel* $k$ *and parameter* $C$*, wrt the* $\|\cdot\|_{\infty;\mathcal{M}}$*-norm.*

*Proof.* Lemma's 12 and 10 combined via the triangle inequality, with Lemma 13, together establish the result as follows. Define $A$ to be the conditioning event regarding the approximation of $k$ by $\hat{k}$, denote the events in Lemma's 12 and 8 by $B$ and $C$ (beware we are overloading $C$ with the regularization parameter; its meaning will be apparent from the context), and the target event in the theorem by $D$.

$$A=\left\{\left\|\hat{k}-k\right\|_{\infty;\mathcal{M}}<\min\left\{1,\frac{\epsilon^2}{2^2\left(\Lambda+2\sqrt{(CL+\frac{\Lambda}{2})\Lambda}\right)^2}\right\}\right\}$$
$$B=\left\{\left\|f^\star-\tilde{f}\right\|_{\infty;\mathcal{M}}\leq\epsilon/2\right\}\qquad C=\left\{\left\|\hat{f}^\star-\tilde{f}\right\|_{\infty}\leq\epsilon/2\right\}\qquad D=\left\{\left\|f^\star-\hat{f}^\star\right\|_{\infty;\mathcal{M}}\leq\epsilon\right\}$$

The claim is a bound on $\Pr(D)$. By the triangle inequality events $B$ and $C$ together imply $D$. Second note that event $C$ is independent of $A$ and $B$. Thus $\Pr(D\mid A)\geq\Pr(B\cap C\mid A)=\Pr(B\mid A)\Pr(C)\geq 1\cdot(1-\delta/2)$, for sufficiently small $\lambda$. Finally Lemma 13 bounds $\Pr(A)$ as follows: provided that $\hat{d}\geq 4(d+2)\log_e\left(2^9\left(\sigma_p\mathrm{diam}(\mathcal{M})\right)^2/\left(\delta\theta(\epsilon)\right)\right)/\theta(\epsilon)$ where $\theta(\epsilon)=\min\left\{1,\epsilon^4/\left[2\left(\Lambda+2\sqrt{(CL+\Lambda/2)\Lambda}\right)\right]^4\right\}$ we have $\Pr(A)\geq 1-\delta/2$. Together this yields $\Pr(D)=\Pr(D\mid A)\Pr(A)\geq(1-\delta/2)^2\geq 1-\delta$. □

Again we see that utility and privacy place competing constraints on the level of noise $\lambda$. Next we will use these interactions to upper-bound the optimal differential privacy of the SVM.

## 5. Hinge-Loss and an Upper Bound on Optimal Differential Privacy

We begin by 'plugging' hinge loss $\ell(y,\hat{y})=(1-y\hat{y})_+$ into the main results on privacy and utility of the previous section (similar computations can be done for PrivateSVM-Finite and other convex loss functions). The following is the dual formulation of hinge-loss SVM learning:

$$\max_{\boldsymbol{\alpha}\in\mathbb{R}^n}\quad\sum_{i=1}^n\alpha_i-\frac{1}{2}\sum_{i=1}^n\sum_{j=1}^n\alpha_i\alpha_jy_iy_jk(\mathbf{x}_i,\mathbf{x}_j)\qquad(5.1)$$
$$\text{s.t.}\quad 0\leq\alpha_i\leq\frac{C}{n}\ \forall i\in[n]\ .$$

**Corollary 15.** *Consider any database* $D$ *of size* $n$*, scalar* $C>0$*, and translation-invariant kernel* $k$*. For any* $\beta>0$ *and* $\hat{d}\in\mathbb{N}$*, PrivateSVM run on* $D$ *with hinge loss, noise parameter* $\lambda\geq\frac{2^{2.5}C\sqrt{\hat{d}}}{\beta n}$*, approximation parameter* $\hat{d}$*, and regularization parameter* $C$*, guarantees* $\beta$*-differential privacy. Moreover for any compact set* $\mathcal{M}\subset\mathbb{R}^d$ *containing* $D$*, and scalars* $\epsilon>0$ *and* $\delta\in(0,1)$*, PrivateSVM run on* $D$ *with hinge loss, kernel* $k$*, noise parameter* $\lambda\leq\min\left\{\frac{\epsilon}{2^4\log_e 2\sqrt{\hat{d}}},\frac{\epsilon\sqrt{\hat{d}}}{8\log_e\frac{2}{\delta}}\right\}$*, approximation parameter* $\hat{d}\geq\frac{4(d+2)}{\theta(\epsilon)}\log_e\left(\frac{2^9(\sigma_p\mathrm{diam}(\mathcal{M}))^2}{\delta\theta(\epsilon)}\right)$ *with* $\theta(\epsilon)=\min\left\{1,\frac{\epsilon^4}{2^{12}C^4}\right\}$*, and regularization parameter* $C$*, is* $(\epsilon,\delta)$*-useful wrt hinge-loss* SVM *run on* $D$ *with kernel* $k$*, and parameter* $C$*.*

*Proof.* The first result follows from Theorem 7 and the fact that hinge-loss is convex and 1-Lipschitz on $\mathbb{R}$: *i.e.*, $\partial_{\hat{y}}\ell = \mathbf{1}[1 \geq y\hat{y}] \leq 1$. The second result follows almost immediately from Theorem 14. For hinge-loss we have that feasible $\alpha_i$'s are bounded by $C/n$ (and so $\Lambda = C$) by the dual's box constraints and that $L = 1$, implying we take $\theta(\epsilon) = \min\left\{1, \frac{\epsilon^4}{2^4 C^4 \left(1+\sqrt{6}\right)^4}\right\}$. This is bounded by the stated $\theta(\epsilon)$. □

Combining the competing requirements on noise level $\lambda$ upper-bounds optimal differential privacy of hinge-loss SVM.

**Theorem 16.** *The optimal differential privacy for hinge-loss* SVM *learning on translation-invariant kernel* $k$ *is bounded by* $\beta(\epsilon, \delta, C, n, \ell, k) = O\left(\frac{1}{\epsilon^3 n}\sqrt{\log\frac{1}{\delta\epsilon}}\left(\log\frac{1}{\epsilon} + \log^2\frac{1}{\delta\epsilon}\right)\right)$.

*Proof.* Consider hinge loss in Corollary 15. Privacy places a lower bound of $\beta \geq 2^{2.5} C\sqrt{\hat{d}}/(\lambda n)$ for any chosen $\lambda$, which we can convert to a lower bound on $\beta$ in terms of $\epsilon$ and $\delta$ as follows. For small $\epsilon$, we have $\theta(\epsilon) = \epsilon^4 2^{-12} C^{-4}$ and so to achieve $(\epsilon, \delta)$-usefulness we must take $\hat{d} = O\left(\frac{1}{\epsilon^4}\log_e\left(\frac{1}{\delta\epsilon^4}\right)\right)$. There are two cases for utility, if $\lambda = \epsilon / \left(2^4 \log_e\left(2\sqrt{\hat{d}}\right)\right)$ then $\beta = O\left(\frac{\sqrt{\hat{d}}\log_e\sqrt{\hat{d}}}{\epsilon n}\right) = O\left(\frac{1}{\epsilon^3 n}\sqrt{\log\frac{1}{\delta\epsilon}}\left(\log\frac{1}{\epsilon} + \log^2\frac{1}{\delta\epsilon}\right)\right)$. Otherwise we are in the second case, with $\lambda = \frac{\epsilon\sqrt{\hat{d}}}{8\log_e\frac{2}{\delta}}$ yielding $\beta = O\left(\frac{1}{\epsilon n}\log\frac{1}{\delta}\right)$ which is dominated by the first case as $\epsilon \downarrow 0$. □

A natural question arises from this discussion: given any mechanism that is $(\epsilon, \delta)$-useful with respect to hinge SVM, for how small a $\beta$ can we possibly hope to guarantee $\beta$-differential privacy? In other words, what lower bounds exist for the optimal differential privacy for the SVM?

## 6. Lower Bounding Optimal Differential Privacy

To lower bound $\beta$ for any $(\epsilon, \delta)$-useful mechanism, we first establish a negative sensitivity result for the SVM, by constructing two neighboring databases on which SVM classifiers differ.

**Lemma 17.** *For any* $C > 0$, $n > 1$ *and* $0 < \epsilon < \frac{\sqrt{C}}{2n}$, *there exists a pair of neighboring databases* $D_1, D_2$ *on* $n$ *entries, such that the functions* $f_1^\star, f_2^\star$ *parametrized by SVM run with parameter* $C$, *linear kernel, and hinge loss on* $D_1, D_2$ *respectively, satisfy* $\|f_1^\star - f_2^\star\|_\infty > 2\epsilon$.

*Proof.* We construct the two databases on the line as follows. Let $0 < m < M$ be scalars to be chosen later. Both databases share negative examples $x_1 = \ldots = x_{\lfloor n/2 \rfloor} = -M$ and positive examples $x_{\lfloor n/2 \rfloor + 1} = \ldots = x_{n-1} = M$. Each database has $x_n = M - m$, with $y_n = -1$ for $D_1$ and $y_n = 1$ for $D_2$. In what follows we use subscripts to denote an example's parent database, so $(x_{i,j}, y_{i,j})$ is the $j^{th}$ example from $D_i$. Consider the result of running primal SVM on each database

$$\begin{aligned} w_1^\star &= \arg\min_{w \in \mathbb{R}} \frac{1}{2}w^2 + \frac{C}{n}\sum_{i=1}^{n}\left(1 - y_{1,i} w x_{1,i}\right)_+ \\ w_2^\star &= \arg\min_{w \in \mathbb{R}} \frac{1}{2}w^2 + \frac{C}{n}\sum_{i=1}^{n}\left(1 - y_{2,i} w x_{2,i}\right)_+ . \end{aligned}$$

Each optimization is strictly convex and unconstrained, so the optimizing $w_1^\star, w_2^\star$ are characterized by the first-order KKT conditions $0 \in \partial_w f_i(w)$ for $f_i$ being the objective function for learning on $D_i$, and $\partial_w$ denoting the subdifferential operator. Now for each $i \in [2]$

$$\partial_w f_i(w) = w - \frac{C}{n}\sum_{j=1}^{n} y_{i,j} x_{i,j} \tilde{\mathbf{1}}\left[1 - y_{i,j} w x_{i,j}\right] ,$$

where

$$\tilde{\mathbf{1}}[x] = \begin{cases} \{0\} , & \text{if } x < 0 \\ [0, 1] , & \text{if } x = 0 \\ \{1\} , & \text{if } x > 0 \end{cases}$$

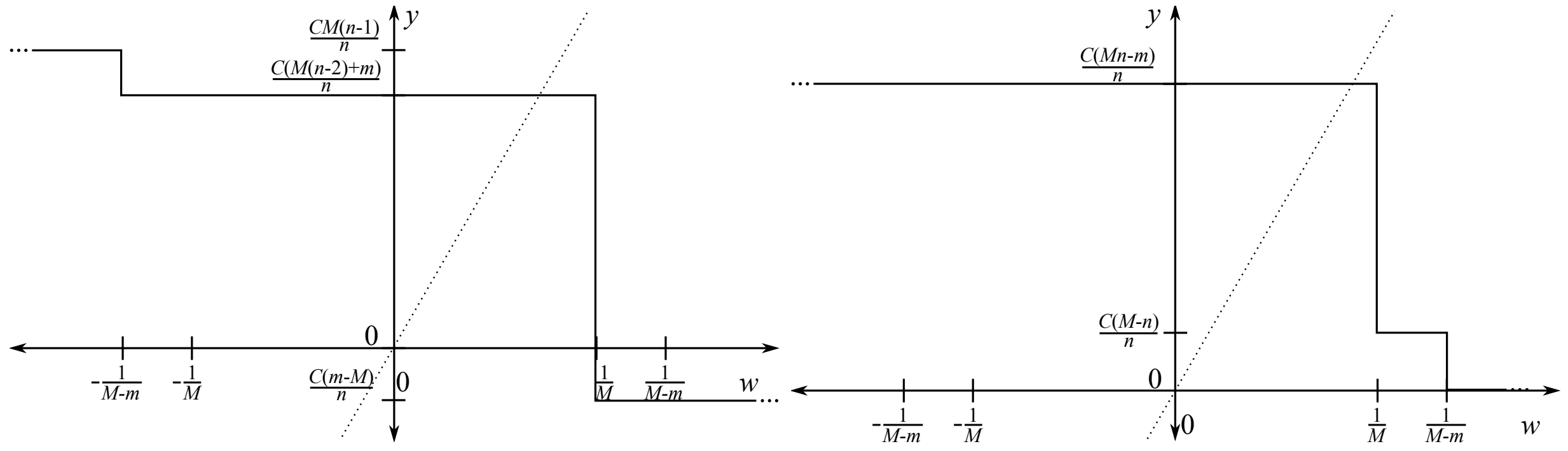

FIGURE 6.1. For each $i \in [2]$, the SVM's primal solution $w_i^\star$on database $D_i$ constructed in the proof of Lemma 17, corresponds to the crossing point of line $y = w$ with $y = w - \partial_w f_i(w)$. Database $D_1$ is shown on the left, database $D_2$ is shown on the right.

is the subdifferential of $(x)_+$. Thus for each $i \in [2]$, $w_i^\star \in \frac{C}{n}\sum_{j=1}^n y_{i,j} x_{i,j} \tilde{\mathbf{1}}\left[1 - y_{i,j} w_i^\star x_{i,j}\right]$ which is equivalent to

$$\begin{aligned} w_1^\star &\in \frac{CM(n-1)}{n}\tilde{\mathbf{1}}\left[\frac{1}{M} - w_1^\star\right] + \frac{C(m-M)}{n}\tilde{\mathbf{1}}\left[w_1^\star - \frac{1}{m-M}\right] \\ w_2^\star &\in \frac{CM(n-1)}{n}\tilde{\mathbf{1}}\left[\frac{1}{M} - w_2^\star\right] + \frac{C(M-m)}{n}\tilde{\mathbf{1}}\left[\frac{1}{M-m} - w_2^\star\right] \ . \end{aligned}$$

The RHSs of these conditions correspond to decreasing piecewise-constant functions, and the conditions are met when the corresponding functions intersect with the diagonal $y = x$ line, as shown in Figure 6. If $\frac{C(M(n-2)+m)}{n} < \frac{1}{M}$ then $w_1^\star = \frac{C(M(n-2)+m)}{n}$. And if $\frac{C(Mn-m)}{n} < \frac{1}{M}$ then $w_2^\star = \frac{C(Mn-m)}{n}$. So provided that $\frac{1}{M} > \frac{C(Mn-m)}{n} = \max\left\{\frac{C(M(n-2)+m)}{n}, \frac{C(Mn-m)}{n}\right\}$, we have $|w_1^\star - w_2^\star| = \frac{2C}{n}|M - m|$. So taking $M = \frac{2n\epsilon}{C}$ and $m = \frac{n\epsilon}{C}$, this implies

$$\begin{aligned} \|f_1^\star - f_2^\star\|_\infty &\geq |f_1^\star(1) - f_2^\star(1)| \\ &= |w_1^\star - w_2^\star| \\ &= 2\epsilon \ , \end{aligned}$$

provided $\epsilon < \frac{\sqrt{C}}{2n}$. □

**Theorem 18** (Lower bound on optimal differential privacy for hinge loss SVM)**.** *For any $C > 0$, $n > 1$, $\delta \in (0, 1)$ and $\epsilon \in \left(0, \frac{\sqrt{C}}{2n}\right)$, the optimal differential privacy for the hinge-loss* SVM *with linear kernel is lower-bounded by $\log_e \frac{1-\delta}{\delta}$. In other words, for any $C, \beta > 0$ and $n > 1$ if a mechanism $\hat{M}$ is $(\epsilon, \delta)$-useful and $\beta$-differentially private then either $\epsilon \geq \frac{\sqrt{C}}{2n}$ or $\delta \geq \exp(-\beta)$.*

*Proof.* Consider $(\epsilon, \delta)$-useful mechanism $\hat{M}$ with respect to SVM learning mechanism $M$ with parameter $C > 0$, hinge loss and linear kernel on $n$ training examples, where $\delta > 0$ and $\frac{\sqrt{C}}{2n} > \epsilon > 0$. By Lemma 17 there exists a pair of neighboring databases $D_1, D_2$ on $n$ entries, such that $\|f_1^\star - f_2^\star\|_\infty > 2\epsilon$ where $f_i^\star = f_{M(D_i)}$ for each $i \in [2]$. Let $\hat{f}_i = f_{\hat{M}(D_i)}$ for each $i \in [2]$. Then by the utility of $\hat{M}$,

$$\Pr\left(\hat{f}_1 \in \mathcal{B}_\epsilon^\infty\left(f_1^\star\right)\right) \geq 1 - \delta \ , \tag{6.1}$$

$$\Pr\left(\hat{f}_2 \in \mathcal{B}_\epsilon^\infty\left(f_1^\star\right)\right) \leq \Pr\left(\hat{f}_2 \notin \mathcal{B}_\epsilon^\infty\left(f_2^\star\right)\right) < \delta \ . \tag{6.2}$$

Let $\hat{\mathcal{P}}_1$ and $\hat{\mathcal{P}}_2$ be the distributions of $\hat{M}(D_1)$ and $\hat{M}(D_2)$ respectively so that $\hat{\mathcal{P}}_i(t) = \Pr\left(\hat{M}(D_i) = t\right)$. Then by Inequalities (6.1) and (6.2)

$$\mathbb{E}_{T\sim\mathcal{P}_1}\left[\frac{d\mathcal{P}_2(T)}{d\mathcal{P}_1(T)} \,\middle|\, T \in \mathcal{B}_\epsilon^\infty\left(f_1^\star\right)\right] = \frac{\int_{\mathcal{B}_\epsilon^\infty(f_1^\star)} \frac{d\mathcal{P}_2(t)}{d\mathcal{P}_1(t)} d\mathcal{P}_1(t)}{\int_{\mathcal{B}_\epsilon^\infty(f_1^\star)} d\mathcal{P}_1(t)} \leq \frac{\delta}{1-\delta} \ .$$

Thus there exists a $t$ such that $\log \frac{\Pr\left(\hat{M}(D_1)=t\right)}{\Pr\left(\hat{M}(D_2)=t\right)} \geq \log \frac{1-\delta}{\delta}$. □

The same technique can be extended to prove a stronger lower bound. First we construct a set of $N > 1$ neighboring databases having SVM images that are a $2\epsilon$-packing. To achieve this for any $N$ we move from linear to RBF kernel.

**Lemma 19.** *For any $C > 0$, $n > C$, $0 < \epsilon < \frac{C}{4n}$, and $0 < \sigma < \sqrt{\frac{1}{2\log_e 2}}$ there exists a set of $N = \left\lfloor \frac{2}{\sigma}\sqrt{\frac{2}{\log_e 2}} \right\rfloor$ pairwise-neighboring databases $\{D_i\}_{i=1}^N$ on $n$ examples, such that the functions $f_i^\star$ parametrized by hinge-loss* SVM *run on $D_i$ with parameter $C$ and RBF kernel with parameter $\sigma$, satisfy $\left\|f_i^\star - f_j^\star\right\|_\infty > 2\epsilon$ for each $i \neq j$.*

*Proof.* Construct $N > 1$ pairwise neighboring databases each on $n$ examples in $\mathbb{R}^2$ as follows. Each database $i$ has $n-1$ negative examples $\mathbf{x}_{i,1} = \ldots = \mathbf{x}_{i,n-1} = \mathbf{0}$, and database $D_i$ has positive example $\mathbf{x}_{i,n} = (\cos\theta_i, \sin\theta_i)$ where $\theta_i = \frac{2\pi i}{N}$. Consider the result of running SVM with hinge loss and RBF kernel on each $D_i$. For each database $k(\mathbf{x}_{i,s}, \mathbf{x}_{i,t}) = 1$ and $k(\mathbf{x}_{i,s}, \mathbf{x}_{i,n}) = \exp\left(-\frac{1}{2\sigma^2}\right) =: \gamma$ for all $s, t \in [n-1]$ . Notice that the range space of $\gamma$ is $(0,1)$. Since the inner-products and labels are database-independent, the SVM dual variables are also database-independent. Each involves solving

$$\begin{aligned} \max_{\boldsymbol{\alpha}\in\mathbb{R}^n} \quad & \boldsymbol{\alpha}'\mathbf{1} - \frac{1}{2}\boldsymbol{\alpha}' \begin{pmatrix} 1 & -\gamma \\ -\gamma & 1 \end{pmatrix} \boldsymbol{\alpha} \\ \text{s.t.} \quad & \mathbf{0} \leq \boldsymbol{\alpha} \leq \frac{C}{n}\mathbf{1} \end{aligned}$$

By symmetry $\alpha_1^\star = \ldots = \alpha_{n-1}^\star$, so we can reduce this to the equivalent program on two variables:

$$\begin{aligned} \max_{\boldsymbol{\alpha}\in\mathbb{R}^2} \quad & \boldsymbol{\alpha}' \begin{pmatrix} n-1 \\ 1 \end{pmatrix} - \frac{1}{2}\boldsymbol{\alpha}' \begin{pmatrix} (n-1)^2 & -\gamma(n-1) \\ -\gamma(n-1) & 1 \end{pmatrix} \boldsymbol{\alpha} \\ \text{s.t.} \quad & \mathbf{0} \leq \boldsymbol{\alpha} \leq \frac{C}{n}\mathbf{1} \end{aligned}$$

Consider first the unconstrained program. In this case the necessary first-order KKT condition is that

$$\mathbf{0} = \begin{pmatrix} n-1 \\ 1 \end{pmatrix} - \begin{pmatrix} (n-1)^2 & -\gamma(n-1) \\ -\gamma(n-1) & 1 \end{pmatrix} \boldsymbol{\alpha}^\star .$$

This implies

$$\begin{aligned} \boldsymbol{\alpha}^\star &= \begin{pmatrix} (n-1)^2 & -\gamma(n-1) \\ -\gamma(n-1) & 1 \end{pmatrix}^{-1} \begin{pmatrix} n-1 \\ 1 \end{pmatrix} \\ &= \frac{1}{(n-1)^2(1-\gamma^2)} \begin{pmatrix} 1 & \gamma(n-1) \\ \gamma(n-1) & (n-1)^2 \end{pmatrix} \begin{pmatrix} n-1 \\ 1 \end{pmatrix} \\ &= \frac{1}{(n-1)^2(1-\gamma)(1+\gamma)} \begin{pmatrix} 1 & \gamma(n-1) \\ \gamma(n-1) & (n-1)^2 \end{pmatrix} \begin{pmatrix} n-1 \\ 1 \end{pmatrix} \\ &= \frac{1}{(n-1)^2(1-\gamma)(1+\gamma)} \begin{pmatrix} (n-1)(1+\gamma) \\ (n-1)^2(1+\gamma) \end{pmatrix} \\ &= \begin{pmatrix} \frac{1}{(n-1)(1-\gamma)} \\ \frac{1}{1-\gamma} \end{pmatrix} . \end{aligned}$$

Since this solution is strictly positive, it follows that at most two (upper) constraints can be active. Thus four cases are possible: the solution lies in the interior of the feasible set, or one or both upper box-constraints hold with equality. Noting that $\frac{1}{(n-1)(1-\gamma)} \leq \frac{1}{1-\gamma}$ it follows that $\boldsymbol{\alpha}^\star$ is feasible iff $\frac{1}{1-\gamma} \leq \frac{C}{n}$. This is equivalent to $C \geq \frac{1}{1-\gamma}n > n$, since $\gamma \in (0,1)$. This corresponds to under-regularization.

If both constraints hold with equality we have $\boldsymbol{\alpha}^\star = \frac{C}{n}\mathbf{1}$, which is always feasible.

In the case where the first constraint holds with equality $\alpha_1^\star = \frac{C}{n}$, the second dual variable is found by optimizing

$$\begin{aligned}
\alpha_2^\star &= \max_{\alpha_2\in\mathbb{R}} \boldsymbol{\alpha}'\begin{pmatrix} n-1 \\ 1 \end{pmatrix} - \frac{1}{2}\boldsymbol{\alpha}'\begin{pmatrix} (n-1)^2 & -\gamma(n-1) \\ -\gamma(n-1) & 1 \end{pmatrix}\boldsymbol{\alpha} \\
&= \max_{\alpha_2\in\mathbb{R}} \frac{C(n-1)}{n} + \alpha_2 - \frac{1}{2}\left(\left(\frac{C(n-1)}{n}\right)^2 - 2\frac{C\gamma(n-1)}{n}\alpha_2 + \alpha_2^2\right) \\
&= \max_{\alpha_2\in\mathbb{R}} -\frac{1}{2}\alpha_2^2 + \alpha_2\left(1 + \frac{C\gamma(n-1)}{n}\right) ,
\end{aligned}$$

implying $\alpha_2^\star = 1 + C\gamma\frac{n-1}{n}$. This solution is feasible provided $1 + C\gamma\frac{n-1}{n} \le \frac{C}{n}$ iff $n \le \frac{C(1+\gamma)}{1+C\gamma}$. Again this corresponds to under-regularization.

Finally in the case where the second constraint holds with equality $\alpha_2^\star = \frac{C}{n}$, the first dual is found by optimizing

$$\begin{aligned}
\alpha_2^\star &= \max_{\alpha_1\in\mathbb{R}} \boldsymbol{\alpha}'\begin{pmatrix} n-1 \\ 1 \end{pmatrix} - \frac{1}{2}\boldsymbol{\alpha}'\begin{pmatrix} (n-1)^2 & -\gamma(n-1) \\ -\gamma(n-1) & 1 \end{pmatrix}\boldsymbol{\alpha} \\
&= \max_{\alpha_1\in\mathbb{R}} (n-1)\alpha_1 + \frac{C}{n} - \frac{1}{2}\left((n-1)^2\alpha_1^2 - 2C\gamma\frac{n-1}{n}\alpha_1 + \frac{C^2}{n^2}\right) \\
&= \max_{\alpha_2\in\mathbb{R}} -\frac{1}{2}(n-1)^2\alpha_1^2 + \alpha_1\left(1 + \frac{C\gamma}{n}\right) ,
\end{aligned}$$

implying $\alpha_1^\star = \frac{1+\frac{C\gamma}{n}}{(n-1)^2}$. This is feasible provided $\frac{1+\frac{C\gamma}{n}}{(n-1)^2} \le \frac{C}{n}$. Passing back to the program on $n$ variables, by the invariance of the duals to the database, for any pair $D_i, D_j$

$$\begin{aligned}
\left|f_i\left(\mathbf{x}_{i,n}\right) - f_j\left(\mathbf{x}_{i,n}\right)\right| &= \alpha_n^\star\left(1 - k\left(\mathbf{x}_{i,n}, \mathbf{x}_{j,n}\right)\right) \\
&\ge \alpha_n^\star\left(1 - \max_{q\ne i} k\left(\mathbf{x}_{i,n}, \mathbf{x}_{q,n}\right)\right) .
\end{aligned}$$

Now a simple argument shows that this maximum is equal to $\gamma^4\exp\left(\sin^2\frac{\pi}{N}\right)$ for all $i$. The maximum objective is optimized when $|q-i| = 1$. In this case $|\theta_i - \theta_q| = \frac{2\pi}{N}$. The norm $\|\mathbf{x}_{i,n} - \mathbf{x}_{q,n}\| = 2\sin\frac{|\theta_i-\theta_q|}{2} = 2\sin\frac{\pi}{N}$ by basic geometry. Thus $k\left(\mathbf{x}_{i,n}, \mathbf{x}_{q,n}\right) = \exp\left(-\frac{\|\mathbf{x}_{i,n}-\mathbf{x}_{q,n}\|^2}{2\sigma^2}\right) = \exp\left(-\frac{2}{\sigma^2}\sin^2\frac{\pi}{N}\right) = \gamma^4\exp\left(\sin^2\frac{\pi}{N}\right)$ as claimed. Notice that $N \ge 2$ so the second term is in $(1, e]$, while the first term is in $(0, 1)$. In summary we have shown that for any $i \ne j$

$$\left|f_i\left(\mathbf{x}_{i,n}\right) - f_j\left(\mathbf{x}_{i,n}\right)\right| \ge \left(1 - \exp\left(-\frac{2}{\sigma^2}\sin^2\frac{\pi}{N}\right)\right)\alpha_n^\star .$$

Assume $\gamma < \frac{1}{2}$. If $n > C$ then $n > \frac{C}{2} > (1-\gamma)C$ in which implies case 1 is infeasible. Similarly since $C\gamma\frac{n-1}{n} > 0$, $n > C$ implies $1 + C\gamma\frac{n-1}{n} > 1 > \frac{C}{n}$ which implies case 3 is infeasible. Thus provided that $\gamma < \frac{1}{2}$ and $n > C$ we have that either case 2 or case 4 must hold. In both cases $\alpha_n^\star = \frac{C}{n}$ giving

$$\left|f_i\left(\mathbf{x}_{i,n}\right) - f_j\left(\mathbf{x}_{i,n}\right)\right| \ge \left(1 - \exp\left(-\frac{2}{\sigma^2}\sin^2\frac{\pi}{N}\right)\right)\frac{C}{n} .$$

Provided that $\sigma \le \sqrt{\frac{2}{\log 2}}\sin\frac{\pi}{N}$ we have $\left(1 - \exp\left(-\frac{2}{\sigma^2}\sin^2\frac{\pi}{N}\right)\right)\frac{C}{n} \ge \left(1 - \frac{1}{2}\right)\frac{C}{n} = \frac{C}{2n}$. Now for small $x$ we can take the linear approximation $\sin x \ge \frac{x}{\pi/2}$ for $x \in [0, \pi/2]$. If $N \ge 2$ then $\sin\frac{\pi}{N} \ge \frac{2}{N}$. Thus in this case we can take $\sigma \le \sqrt{\frac{2}{\log 2}}\frac{2}{N}$ to imply $\left|f_i\left(\mathbf{x}_{i,n}\right) - f_j\left(\mathbf{x}_{i,n}\right)\right| \ge \frac{C}{2n}$. This bound on $\sigma$ in turn implies the following bound on $\gamma$: $\gamma = \exp\left(-\frac{1}{2\sigma^2}\right) \le \exp\left(-\frac{N^2\log_e 2}{2^4}\right)$. Thus taking $N > 4$, in conjunction with $\sigma \le \sqrt{\frac{2}{\log 2}}\frac{2}{N}$ implies $\gamma \le \frac{1}{2}$. Rather than selecting $N$ which bounds $\sigma$, we can choose $N$ in terms of $\sigma$. $\sigma \le \sqrt{\frac{2}{\log 2}}\frac{2}{N}$ is implied by $N = \frac{2}{\sigma}\sqrt{\frac{2}{\log_e 2}}$. So for small $\sigma$ we can construct more databases leading to the desired separation. Finally, $N > 4$ implies that we must constrain $\sigma < \sqrt{\frac{1}{2\log_e 2}}$.

In summary, if $n > C$ and $\sigma < \sqrt{\frac{1}{2\log_e 2}}$ then $|f_i(\mathbf{x}_{i,n}) - f_j(\mathbf{x}_{i,n})| \geq \frac{C}{2n}$ for each $i \neq j \in [N]$ where $N = \left\lfloor \frac{2}{\sigma}\sqrt{\frac{2}{\log_e 2}} \right\rfloor$. Moreover if $\epsilon \leq \frac{C}{4n}$ then for any $i \neq j$ this implies $\|f_i - f_j\|_\infty \geq 2\epsilon$ as claimed. □

**Theorem 20** (Strong lower bound on optimal differential privacy for hinge loss). *For $C > 0$, $n > C$, $\delta \in (0,1)$, $\epsilon \in \left(0, \frac{n}{4C}\right)$, and $\sigma < \sqrt{\frac{1}{2\log_e 2}}$ the optimal differential privacy for the hinge* SVM *with RBF kernel having parameter $\sigma$ is lower-bounded by $\log_e \frac{(1-\delta)(N-1)}{\delta}$, where $N = \left\lfloor \frac{2}{\sigma}\sqrt{\frac{2}{\log_e 2}} \right\rfloor$. That is, under these conditions, all mechanisms that are $(\epsilon,\delta)$-useful wrt hinge* SVM *with RBF kernel for any $\sigma$ do not achieve differential privacy at any level.*

*Proof.* Consider $(\epsilon,\delta)$-useful mechanism $\hat{M}$ with respect to hinge SVM learning mechanism $M$ with parameter $C > 0$ and RBF kernel with parameter $0 < \sigma < \sqrt{\frac{1}{2\log_e 2}}$ on $n$ training examples, where $\delta > 0$ and $\frac{n}{4C} > \epsilon > 0$. Let $N = \left\lfloor \frac{2}{\sigma}\sqrt{\frac{2}{\log_e 2}} \right\rfloor > 4$. By Lemma 19 there exist pairwise neighboring databases $D_1, \ldots, D_N$ of $n$ entries, such that $\{f_i^\star\}_{i=1}^N$ is an $\epsilon$-packing wrt the $L_\infty$-norm, where $f_i^\star = f_{M(D_i)}$. So by the utility of $\hat{M}$, for each $i \in [N]$

$$
\begin{aligned}
\text{(6.3)} \qquad \Pr\left(\hat{f}_i \in \mathcal{B}_\epsilon^\infty(f_i^\star)\right) &\geq 1-\delta\,, \\
\sum_{j\neq 1} \Pr\left(\hat{f}_1 \in \mathcal{B}_\epsilon^\infty(f_j^\star)\right) &\leq \Pr\left(\hat{f}_1 \notin \mathcal{B}_\epsilon^\infty(f_1^\star)\right) < \delta\,, \\
\text{(6.4)} \qquad \Rightarrow \exists j \neq 1,\ \Pr\left(\hat{f}_1 \in \mathcal{B}_\epsilon^\infty(f_j^\star)\right) &< \frac{\delta}{N-1}\,.
\end{aligned}
$$

Let $\hat{\mathcal{P}}_1$ and $\hat{\mathcal{P}}_j$ be the distributions of $\hat{M}(D_1)$ and $\hat{M}(D_j)$ respectively so that for each, $\hat{\mathcal{P}}_i(t) = \Pr\left(\hat{M}(D_i) = t\right)$. Then by Inequalities (6.3) and (6.4)

$$
\mathbb{E}_{T\sim\mathcal{P}_j}\left[\frac{d\mathcal{P}_1(T)}{d\mathcal{P}_j(T)} \,\middle|\, T \in \mathcal{B}_\epsilon^\infty(f_j^\star)\right] = \frac{\int_{\mathcal{B}_\epsilon^\infty(f_j^\star)} \frac{d\mathcal{P}_1(t)}{d\mathcal{P}_j(t)} d\mathcal{P}_j(t)}{\int_{\mathcal{B}_\epsilon^\infty(f_j^\star)} d\mathcal{P}_j(t)} \leq \frac{\delta}{(1-\delta)(N-1)}\,.
$$

Thus there exists a $t$ such that $\log \frac{\Pr\left(\hat{M}(D_j)=t\right)}{\Pr\left(\hat{M}(D_1)=t\right)} \geq \log \frac{(1-\delta)(N-1)}{\delta}$. □

Note that $n > C$ is a weak condition, since $C$ should grow like $\sqrt{n}$ for universal consistency. Also note that this negative result is consistent with our upper bound on optimal differential privacy: $\sigma$ affects $\sigma_p$, increasing the upper bounds as $\sigma \downarrow 0$.

## 7. Conclusion & Open Problems

We have presented a pair of new mechanisms for private SVM learning. In each case we have established differential privacy via the algorithmic stability of regularized empirical risk minimization. To achieve utility under infinite-dimensional feature mappings, we perform regularized ERM in a random Reproducing Kernel Hilbert Space whose kernel approximates the target RKHS kernel. This trick, borrowed from large-scale learning, permits the mechanism to privately respond with a finite representation of a maximum-margin hyperplane classifier. We then established the high-probability, pointwise similarity between the resulting function and the SVM classifier through a new smoothness result of regularized ERM with respect to perturbations of the RKHS. The bounds on differential privacy and utility combine to upper bound the optimal differential privacy of SVM learning for hinge-loss. This quantity is the optimal level of privacy among all mechanisms that are $(\epsilon,\delta)$-useful with respect to the hinge-loss SVM. Finally, we derived a lower bound on this quantity which established that any mechanism that is too accurate with respect to the hinge SVM with RBF kernel, with any non-trivial probability, cannot be $\beta$-differentially private for small $\beta$. The lower bounds explicitly depend on the variance of the RBF kernel.

An interesting open problem is to derive lower bounds holding for moderate to large $\epsilon$. Another direction for future research is to extend our mechanisms to other kernel methods. Finally, a general connection between algorithmic stability and global sensitivity would immediately suggest a number of practical privacy-preserving learning mechanisms.

## Acknowledgments

We thank Ali Rahimi for helpful discussions relating to this research. We gratefully acknowledge the support of the NSF through grant DMS-0707060, and the support of the Siebel Scholars Foundation.

## Appendix A. Proofs for Subdifferentiable Loss Functions

The main results were stated in terms of non-differentiable convex loss functions so that they would hold for the hinge loss, however the proofs in the main text applied to differentiable loss functions only. For completeness we now re-prove the appropriate lemma's for subdifferentiable loss functions of which general convex loss functions are a special case.

In each case the proofs for subdifferentiable loss are essentially identical to the differentiable loss proofs: we discuss only the arguments that change when generalizing to non-differentiable loss functions. Previously $\partial_{\mathbf{w}}$ referred to the gradient operator, now it refers to the subdifferential operator. The subscript reminds us that we are viewing the operand as a function of $\mathbf{w}$ only. Similarly other subscripts extend the notion of other partial derivatives. Previously there was a unique gradient at each point, now there may be many subgradients, making up the subdifferential set at a point. As we are dealing with sets of subgradients, we use the shorthand that for sets $S, T$, vector $\mathbf{v}$ and scalar $a$ that: $S + T = \{g + h \mid g \in S, h \in T\}$, $aS = \{ag \mid g \in S\}$, $S + \mathbf{v} = \{\mathbf{g} + \mathbf{v} \mid \mathbf{g} \in S\}$, $\langle S, \mathbf{v}\rangle = \{\langle \mathbf{g}, \mathbf{v}\rangle \mid \mathbf{g} \in S\}$ and $S \geq T$ means $g \geq h$ for all $(g, h) \in S \times T$.

Lemma 21 generalizes Lemma 6 on the sensitivity of the SVM primal weight vector, to general (*i.e.*, subdifferentiable) convex loss functions.

**Lemma 21.** *Consider loss function $\ell(y, \hat{y})$ that is convex and $L$-Lipschitz in $\hat{y}$, and RKHS $\mathcal{H}$ induced by finite $F$-dimensional feature mapping $\phi$ with bounded kernel $k(\mathbf{x}, \mathbf{x}) \leq \kappa^2$ for all $\mathbf{x} \in \mathbb{R}^d$. Let $\mathbf{w}_S \in \mathbb{R}^F$ be the minimizer of the following regularized empirical risk function for each database $S = \{(\mathbf{x}_i, y_i)\}_{i=1}^n$*

$$R_{\mathrm{reg}}(\mathbf{w}, S) \quad = \quad \frac{C}{n}\sum_{i=1}^{n} \ell\left(y_i, f_{\mathbf{w}}(\mathbf{x}_i)\right) + \frac{1}{2}\|\mathbf{w}\|_2^2 \; .$$

*Then for every pair of neighboring databases $D, D'$ of $n$ entries, $\|\mathbf{w}_D - \mathbf{w}_{D'}\|_1 \leq 4LC\kappa\sqrt{F}/n$.*

*Proof.* For convenience we define $R_{\mathrm{emp}}(\mathbf{w}, S) = n^{-1}\sum_{i=1}^n \ell(y_i, f_{\mathbf{w}}(\mathbf{x}_i)$ for any training set $S$, then the first-order necessary KKT conditions imply

$$\mathbf{0} \;\in\; \partial_{\mathbf{w}} R_{\mathrm{reg}}(\mathbf{w}_D, D) \;=\; C\partial_{\mathbf{w}} R_{\mathrm{emp}}(\mathbf{w}_D, D) + \mathbf{w}_D \; , \tag{A.1}$$

$$\mathbf{0} \;\in\; \partial_{\mathbf{w}} R_{\mathrm{reg}}(\mathbf{w}_{D'}, D') \;=\; C\partial_{\mathbf{w}} R_{\mathrm{emp}}(\mathbf{w}_{D'}, D') + \mathbf{w}_{D'} \; . \tag{A.2}$$

Define the auxiliary risk function

$$\tilde{R}(\mathbf{w}) \quad = \quad C\langle \partial_{\mathbf{w}} R_{\mathrm{emp}}(\mathbf{w}_D, D) - \partial_{\mathbf{w}} R_{\mathrm{emp}}(\mathbf{w}_{D'}, D'),\, \mathbf{w} - \mathbf{w}_{D'}\rangle + \frac{1}{2}\|\mathbf{w} - \mathbf{w}_{D'}\|_2^2 \; .$$

It is easy to see that $\tilde{R}(\mathbf{w})$ is strictly convex in $\mathbf{w}$ and that $\tilde{R}(\mathbf{w}_{D'}) = \{0\}$. And by Equation (A.2)

$$\begin{aligned} C\partial_{\mathbf{w}} R_{\mathrm{emp}}(\mathbf{w}_D, D) + \mathbf{w} \quad &\in \quad C\partial_{\mathbf{w}} R_{\mathrm{emp}}(\mathbf{w}_D, D) - C\partial_{\mathbf{w}} R_{\mathrm{emp}}(\mathbf{w}_{D'}, D') + \mathbf{w} - \mathbf{w}_{D'} \\ &= \quad \partial_{\mathbf{w}} \tilde{R}(\mathbf{w}) \; , \end{aligned}$$

which combined with Equation (A.1) implies $\mathbf{0} \in \partial_{\mathbf{w}} \tilde{R}(\mathbf{w}_D)$, so that $\tilde{R}(\mathbf{w})$ is minimized at $\mathbf{w}_D$. Thus there exists some non-positive $r \in \tilde{R}(\mathbf{w}_D)$. Next simplify the first term of $\tilde{R}(\mathbf{w}_D)$, scaled by $n/C$ for notational

convenience:

$$
\begin{aligned}
& n\langle \partial_{\mathbf{w}} R_{\text{emp}}(\mathbf{w}_D, D) - \partial_{\mathbf{w}} R_{\text{emp}}(\mathbf{w}_{D'}, D'),\, \mathbf{w} - \mathbf{w}_{D'} \rangle \\
= & \sum_{i=1}^{n} \langle \partial_{\mathbf{w}} \ell\left(y_i, f_{\mathbf{w}_D}(\mathbf{x}_i)\right) - \partial_{\mathbf{w}} \ell\left(y_i', f_{\mathbf{w}_{D'}}(\mathbf{x}_i')\right),\, \mathbf{w} - \mathbf{w}_{D'} \rangle \\
= & \sum_{i=1}^{n-1} \left(\ell'\left(y_i, f_{\mathbf{w}_D}(\mathbf{x}_i)\right) - \ell'\left(y_i, f_{\mathbf{w}_{D'}}(\mathbf{x}_i)\right)\right)\left(f_{\mathbf{w}_D}(\mathbf{x}_i) - f_{\mathbf{w}_{D'}}(\mathbf{x}_i)\right) \\
& + \ell'\left(y_n, f_{\mathbf{w}_D}(\mathbf{x}_n)\right)\left(f_{\mathbf{w}_D}(\mathbf{x}_n) - f_{\mathbf{w}_{D'}}(\mathbf{x}_n)\right) - \ell'\left(y_n', f_{\mathbf{w}_{D'}}(\mathbf{x}_n')\right)\left(f_{\mathbf{w}_D}(\mathbf{x}_n') - f_{\mathbf{w}_{D'}}(\mathbf{x}_n')\right) \\
\geq & \ \ell'\left(y_n, f_{\mathbf{w}_D}(\mathbf{x}_n)\right)\left(f_{\mathbf{w}_D}(\mathbf{x}_n) - f_{\mathbf{w}_{D'}}(\mathbf{x}_n)\right) - \ell'\left(y_n', f_{\mathbf{w}_{D'}}(\mathbf{x}_n')\right)\left(f_{\mathbf{w}_D}(\mathbf{x}_n') - f_{\mathbf{w}_{D'}}(\mathbf{x}_n')\right) ,
\end{aligned}
$$

where the second equality follows from $\partial_{\mathbf{w}} \ell\left(y, f_{\mathbf{w}}(\mathbf{x})\right) = \ell'\left(y, f_{\mathbf{w}}(\mathbf{x})\right) \phi(\mathbf{x})$, where $\ell'(y, \hat{y}) = \partial_{\hat{y}} \ell(y, \hat{y})$, and $\mathbf{x}_i' = \mathbf{x}_i$ and $y_i' = y_i$ for each $i \in [n-1]$. The inequality follows from the convexity of $\ell$ in its second argument.[6] Combined with the existence of non-positive $r \in \tilde{R}(\mathbf{w}_D)$ this yields that there exists $g \in \ell'\left(y_n', f_{\mathbf{w}_{D'}}(\mathbf{x}_n')\right)\left(f_{\mathbf{w}_D}(\mathbf{x}_n') - f_{\mathbf{w}_{D'}}(\mathbf{x}_n')\right) - \ell'\left(y_n, f_{\mathbf{w}_D}(\mathbf{x}_n)\right)\left(f_{\mathbf{w}_D}(\mathbf{x}_n) - f_{\mathbf{w}_{D'}}(\mathbf{x}_n)\right)$ such that

$$
\begin{aligned}
0 & \geq \frac{n}{C} r \\
& \geq g + \frac{n}{2C} \|\mathbf{w}_D - \mathbf{w}_{D'}\|_2^2
\end{aligned}
$$

And since $|g| \leq 2L \left\| f_{\mathbf{w}_D} - f_{\mathbf{w}_{D'}} \right\|_\infty$ by the Lipschitz continuity of $\ell$, this in turn implies

$$
\frac{n}{2C} \|\mathbf{w}_D - \mathbf{w}_{D'}\|_2^2 \quad \leq \quad 2L \left\| f_{\mathbf{w}_D} - f_{\mathbf{w}_{D'}} \right\|_\infty . \tag{A.3}
$$

Now by the reproducing property and Cauchy-Schwartz inequality we can upper bound the classifier difference's infinity norm by the Euclidean norm on the weight vectors: for each $\mathbf{x}$

$$
\begin{aligned}
\left| f_{\mathbf{w}_D}(\mathbf{x}) - f_{\mathbf{w}_{D'}}(\mathbf{x}) \right| & = |\langle \phi(\mathbf{x}), \mathbf{w}_D - \mathbf{w}_{D'} \rangle| \\
& \leq \|\phi(\mathbf{x})\|_2 \|\mathbf{w}_D - \mathbf{w}_{D'}\|_2 \\
& = \sqrt{k(\mathbf{x}, \mathbf{x})} \|\mathbf{w}_D - \mathbf{w}_{D'}\|_2 \\
& \leq \kappa \|\mathbf{w}_D - \mathbf{w}_{D'}\|_2 .
\end{aligned}
$$

Combining this with Inequality (A.3) yields $\|\mathbf{w}_D - \mathbf{w}_{D'}\|_2 \leq 4LC\kappa/n$ as claimed. The $L_1$-based sensitivity then follows from $\|\mathbf{w}\|_1 \leq \sqrt{F} \|\mathbf{w}\|_2$ for all $\mathbf{w} \in \mathbb{R}^F$. □

Next we move to proofs of utility. Lemma 22 mirrors Lemma 12, generalizing the result to non-differentiable convex loss functions.

**Lemma 22.** *Let $\mathcal{H}$ be an RKHS with translation-invariant kernel $k$, and let $\hat{\mathcal{H}}$ be the random RKHS corresponding to feature map (4.1) induced by $k$. Let $C$ be a positive scalar and loss $\ell(y, \hat{y})$ be convex and $L$-Lipschitz continuous in $\hat{y}$. Consider the regularized empirical risk minimizers in each RKHS*

$$
\begin{aligned}
f^\star & \in \arg\min_{f \in \mathcal{H}} \frac{C}{n} \sum_{i=1}^{n} \ell(y_i, f(\mathbf{x}_i)) + \frac{1}{2} \|f\|_{\mathcal{H}}^2 , \\
g^\star & \in \arg\min_{g \in \hat{\mathcal{H}}} \frac{C}{n} \sum_{i=1}^{n} \ell(y_i, g(\mathbf{x}_i)) + \frac{1}{2} \|g\|_{\hat{\mathcal{H}}}^2 .
\end{aligned}
$$

*Let $\mathcal{M} \subseteq \mathbb{R}^d$ be any set containing $\mathbf{x}_1, \ldots, \mathbf{x}_n$. For any $\epsilon > 0$, if the dual variables from both optimizations have $L_1$-norms bounded by some $\Lambda > 0$ and $\left\| k - \hat{k} \right\|_{\infty; \mathcal{M}} \leq \min\left\{1, \frac{\epsilon^2}{2^2\left(\Lambda + 2\sqrt{(CL + \Lambda/2)\Lambda}\right)^2}\right\}$ then $\|f^\star - g^\star\|_{\infty; \mathcal{M}} \leq \epsilon/2$.*

[6]Namely for convex $f$ and any $a, b \in \mathbb{R}$, $(g_a - g_b)(a - b) \geq 0$ for all $g_a \in \partial f(a)$ and all $g_b \in \partial f(b)$.

*Proof.* Denote the empirical risk functional $R_{\text{emp}}[f] = n^{-1}\sum_{i=1}^n \ell\left(y_i, f(\mathbf{x}_i)\right)$ and the regularized empirical risk functional $R_{\text{reg}}[f] = C\, R_{\text{emp}}[f] + \|f\|^2/2$, for the appropriate RKHS norm (either $\mathcal{H}$ or $\hat{\mathcal{H}}$). Let $f^\star$ denote the regularized empirical risk minimizer in $\mathcal{H}$, given by parameter vector $\boldsymbol{\alpha}^\star$, and let $g^\star$ denote the regularized empirical risk minimizer in $\hat{\mathcal{H}}$ given by parameter vector $\boldsymbol{\beta}^\star$. Let $g_{\alpha^\star} = \sum_{i=1}^n \alpha_i^\star y_i \hat{\phi}(\mathbf{x}_i)$ and $f_{\boldsymbol{\beta}^\star} = \sum_{i=1}^n \beta_i^\star y_i \phi(\mathbf{x}_i)$ denote the images of $f^\star$ and $g^\star$ under the natural mapping between the spans of the data in RKHS's $\hat{\mathcal{H}}$ and $\mathcal{H}$ respectively. We will first show that these four functions have arbitrarily close regularized empirical risk in their respective RKHS, and then that this implies uniform proximity of the functions themselves. First observe that for any $g \in \hat{\mathcal{H}}$

$$
\begin{aligned}
R_{\text{reg}}^{\hat{\mathcal{H}}}[g] &= C\, R_{\text{emp}}[g] + \frac{1}{2}\|g\|_{\hat{\mathcal{H}}}^2 \\
&\geq C\langle \partial_g R_{\text{emp}}[g^\star], g - g^\star\rangle_{\hat{\mathcal{H}}} + C\, R_{\text{emp}}[g^\star] + \frac{1}{2}\|g\|_{\hat{\mathcal{H}}}^2 \\
&= \langle \partial_g R_{\text{reg}}^{\hat{\mathcal{H}}}[g^\star], g - g^\star\rangle_{\hat{\mathcal{H}}} - \langle g^\star, g - g^\star\rangle_{\hat{\mathcal{H}}} + C\, R_{\text{emp}}[g^\star] + \frac{1}{2}\|g\|_{\hat{\mathcal{H}}}^2 \;.
\end{aligned}
$$

The inequality follows from the convexity of $R_{\text{emp}}[\cdot]$ and holds for all elements of the subdifferential $\partial_g R_{\text{emp}}[g^\star]$. The subsequent equality holds by $\partial_g R_{\text{reg}}^{\hat{\mathcal{H}}}[g] = C\, \partial_g R_{\text{emp}}[g] + g$. Now since $\mathbf{0} \in \partial_g R_{\text{reg}}^{\hat{\mathcal{H}}}[g^\star]$, it follows that

$$
\begin{aligned}
R_{\text{reg}}^{\hat{\mathcal{H}}}[g] &\geq C\, R_{\text{emp}}[g^\star] + \frac{1}{2}\|g\|_{\hat{\mathcal{H}}}^2 - \langle g^\star, g - g^\star\rangle_{\hat{\mathcal{H}}} \\
&= C\, R_{\text{emp}}[g^\star] + \frac{1}{2}\|g^\star\|_{\hat{\mathcal{H}}}^2 + \frac{1}{2}\|g\|_{\hat{\mathcal{H}}}^2 - \frac{1}{2}\|g^\star\|_{\hat{\mathcal{H}}}^2 - \langle g^\star, g - g^\star\rangle_{\hat{\mathcal{H}}} \\
&= R_{\text{reg}}^{\hat{\mathcal{H}}}[g^\star] + \frac{1}{2}\|g\|_{\hat{\mathcal{H}}}^2 - \langle g^\star, g\rangle_{\hat{\mathcal{H}}} + \frac{1}{2}\|g^\star\|_{\hat{\mathcal{H}}}^2 \\
&= R_{\text{reg}}^{\hat{\mathcal{H}}}[g^\star] + \frac{1}{2}\|g - g^\star\|_{\hat{\mathcal{H}}}^2 \;.
\end{aligned}
$$

The remainder of Lemma 12's proof remains the same, as it does not depend on the loss's differentiability. $\square$